\documentclass[10pt,twocolumn,letterpaper]{article}

\usepackage {cvpr}      
\usepackage{multirow}
\usepackage{array}
\usepackage{colortbl}
\usepackage{graphicx}
\usepackage{tabularx}
\usepackage{booktabs}
\usepackage{subcaption}
\usepackage{floatrow}
\definecolor{lightgray}{gray}{0.9}
\newcommand{\ccl}{\cellcolor{lightgray}}
\newfloatcommand{capbtabbox}{table}[][\FBwidth]

\definecolor{cvprblue}{rgb}{0.21,0.49,0.74}
\usepackage[pagebackref,breaklinks,colorlinks,allcolors=cvprblue]{hyperref}

\def\paperID{11165} 
\def\confName{CVPR}
\def\confYear{2025}

\title{O-TPT: Orthogonality Constraints for Calibrating Test-time Prompt Tuning in Vision-Language Models}

\author{
    Ashshak Sharifdeen$^{1,2}$, \hspace{0.25em} Muhammad Akhtar Munir$^{1}$, \hspace{0.25em} Sanoojan Baliah$^{1}$, \hspace{0.25em} Salman Khan$^{1,3}$, \\ \hspace{0.25em} Muhammad Haris Khan$^{1}$ \\
    $^{1}$Mohamed bin Zayed University of AI, $^{2}$University of Colombo, $^{3}$Australian National University \\
    {\tt\small \{ashshak.sharifdeen, muhammad.haris\}@mbzuai.ac.ae}
}

\begin{document}
\maketitle
\begin{abstract}

Test-time prompt tuning for vision-language models (VLMs) is getting attention because of their ability to learn with unlabeled data without fine-tuning. 
Although test-time prompt tuning methods for VLMs can boost accuracy, the resulting models tend to demonstrate poor calibration, which casts doubts on the reliability and trustworthiness of these models. 
Notably, more attention needs to be devoted to calibrating the test-time prompt tuning in vision-language models. 
To this end, we propose a new approach, called \texttt{O-TPT} that introduces orthogonality constraints on the textual features corresponding to the learnable prompts for calibrating test-time prompt tuning in VLMs.
Towards introducing orthogonality constraints, we make the following contributions. 
First, we uncover new insights behind the suboptimal calibration performance of existing methods relying on textual feature dispersion. Second, we show that imposing a simple orthogonalization of textual features is a more effective approach towards obtaining textual dispersion.
We conduct extensive experiments on various datasets with different backbones and baselines. The results indicate that our method consistently outperforms the prior state of the art in significantly reducing the overall average calibration error. Also, our method surpasses the zero-shot calibration performance on fine-grained classification tasks. Our code is available at \url{https://github.com/ashshaksharifdeen/O-TPT}.
\end{abstract}

\vspace{-0.1in}
\section{Introduction}
\label{sec:intro}

In recent years, vision-language models (VLMs), for example, CLIP \cite{radford2021learning}, have shown impressive zero-shot inference abilities in various downstream tasks \cite{radford2021learning, jia2021scaling}. 
VLMs align web-scale image-text data to learn a shared embedding space, enabling it to classify instances from novel categories in a zero-shot manner via carefully designed prompt templates. So, constructing these manually designed prompts is crucial for effective zero-shot transfer. However, such manually crafted prompts depend on domain-specific heuristics, which may limit their effectiveness \cite{shu2022test}. Prompt tuning aims to learn prompts using downstream training data \cite{zhou2022learning, zhou2022conditional}; however, learned prompts struggle to generalize beyond the task and training data. In addition, prompt-tuning methods require labeled data, which can be costly or even unavailable. 

\begin{figure}
    \centering
    \includegraphics[width=\linewidth]{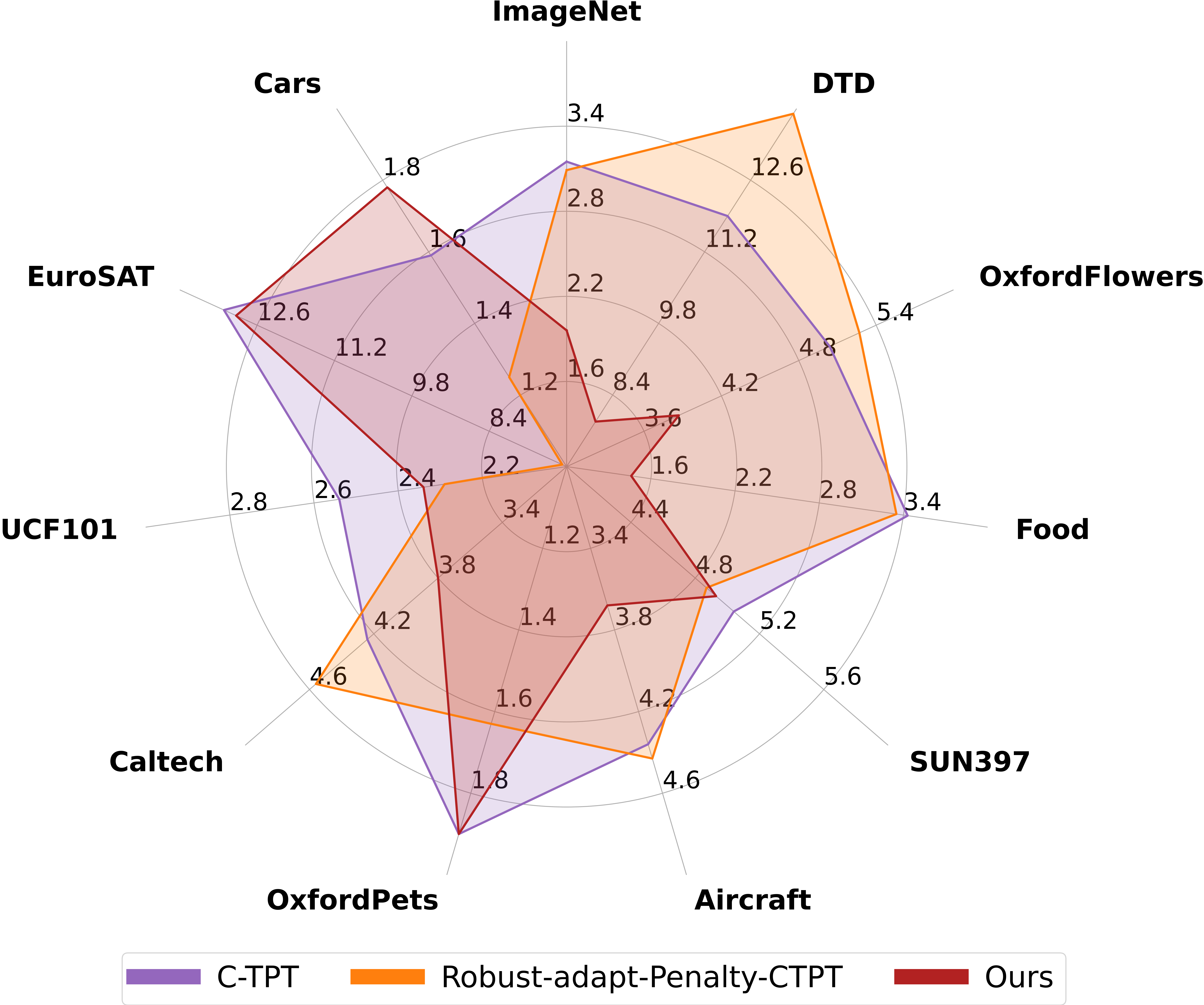}
    \caption{Comparison of calibration performance (ECE) with C-TPT \cite{yoon2024c} and Robust-adapt-Penalty-CTPT\cite{murugesanrobust}. \textbf{Lower} the ECE better the calibration.}
    \label{fig:radar_plot}
\end{figure}

Recently, test-time prompt tuning \cite{shu2022test} has emerged as a paradigm for refining prompts during inference, enabling VLMs to better adapt to newer tasks without labeled data and retraining. 
However, while prompt tuning at the test-time has been shown to enhance the accuracy of the VLM, it usually results in poorly calibrated predictions, which means that the confidence of the model does not align well with its accuracy \cite{murugesanrobust, yoon2024c}. 
This lack of calibration raises concerns about the trustworthiness of VLMs in applications that demand reliable uncertainty estimates, such as healthcare \cite{ji2021improving, wang2022medclip} or autonomous systems \cite{cui2024survey, you2024v2x, zhou2024vision}. 
%
Poorly calibrated predictions can lead to overconfidence, which can lead to undesirable consequences in safety-critical applications, where decisions often depend on model outputs.
We note that calibration in test-time prompt tuning for VLMs remains largely unexplored. 

\begin{figure*}[!t]
    \centering
    \begin{floatrow}
        \ffigbox{%
            \includegraphics[width=\linewidth]{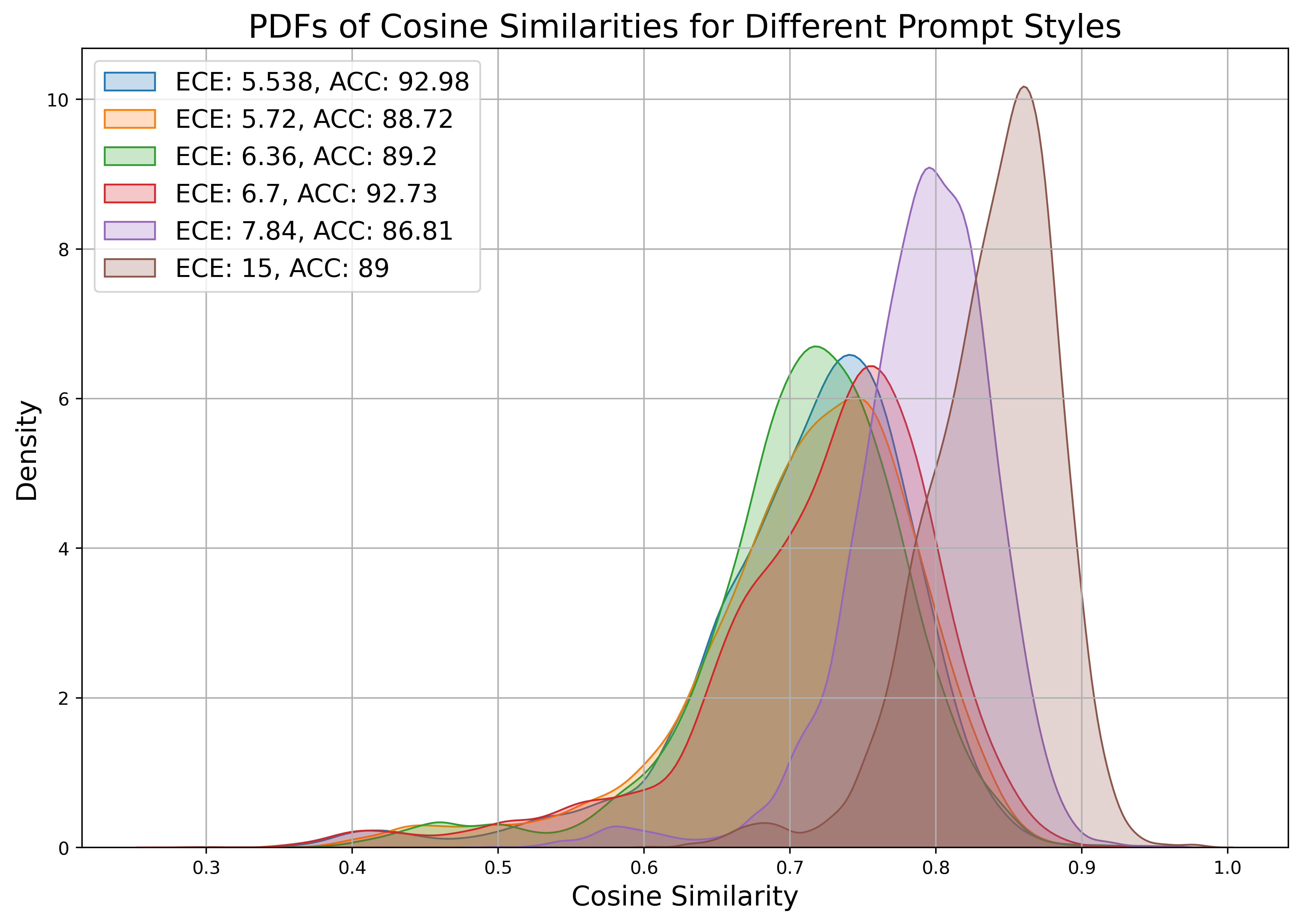}
        }{%
        \vspace{-2.5em}
            \caption{Probability Density Functions of intra-text feature cosine similarities}
            \label{fig:cosine_sim_hard_prompt}
        }
        
        \ffigbox{%
            \includegraphics[width=\linewidth]{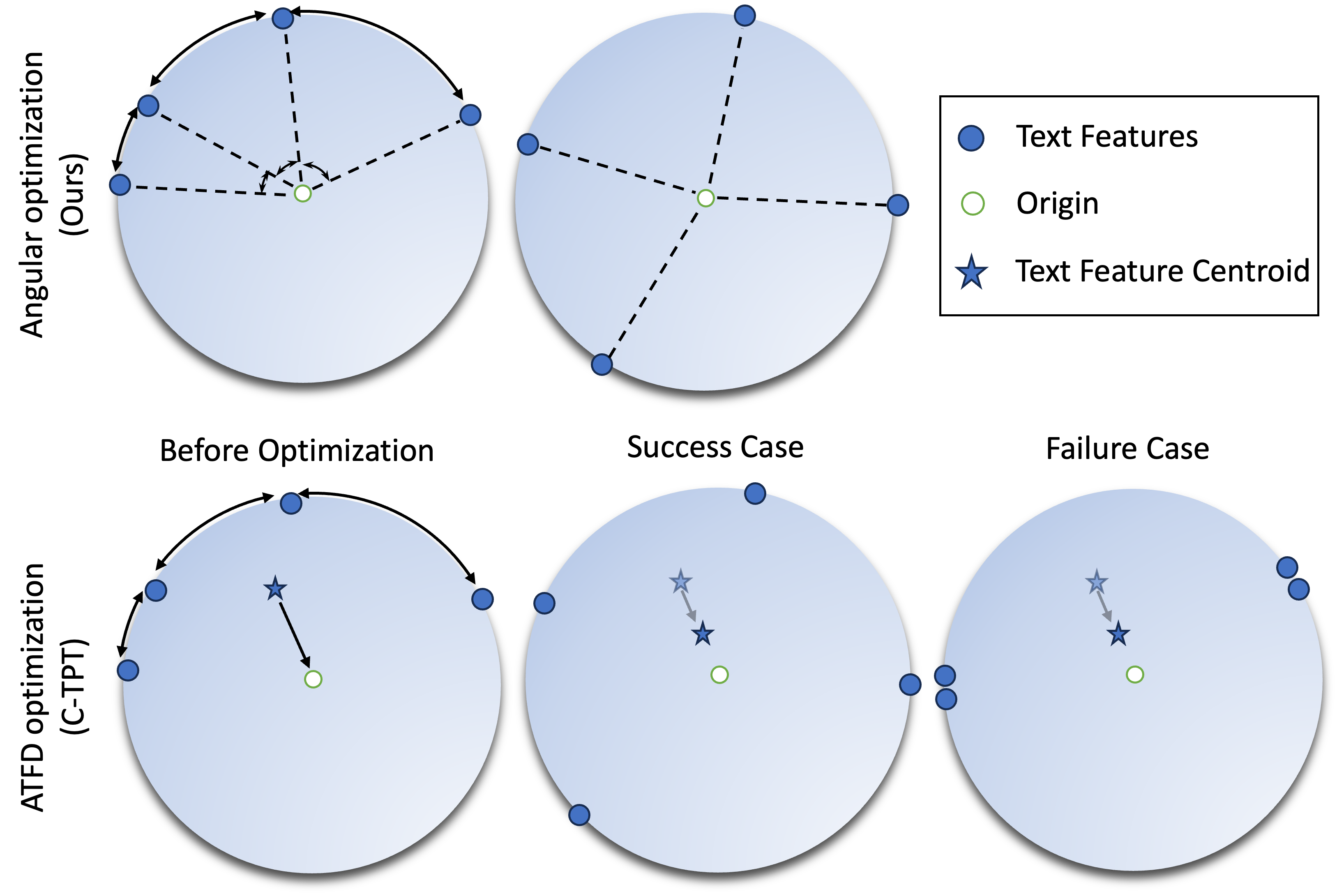}
        }{%
        \vspace{-2.5em}
            \caption{Comparison of angular optimization (ours) and ATFD optimization (C-TPT) \cite{yoon2024c}}
            \label{fig:ctpt_vs_ours_diag}
        
        }
    \end{floatrow}
\end{figure*}

We note that calibration in test-time prompt tuning for VLMs remains largely unexplored with very few attempts. A recent method C-TPT \cite{yoon2024c} observes that test-time prompt tuning methods \cite{shu2022test} often sacrifice the model's calibration performance while boosting its accuracy, leading to unreliable prediction confidence for possibly many test examples. It shows that enhancing text feature dispersion can mitigate the overconfidence issue arising in test-time prompt tuning. However, the method could struggle to ensure the desired dispersion in challenging cases. This can lead to the underutilization of the textual feature space, potentially resulting in textual features lying in close proximity (Fig.~\ref{fig:ctpt_vs_ours_diag}) and eventually causing poor calibration performance (Fig.~\ref{fig:radar_plot}).


We note that the core methodology of C-TPT \cite{yoon2024c} overlooks the critical relationship between the angular separation among textual features and the calibration performance. Our findings reveal a strong correlation: textual features with lower cosine similarity (i.e., greater angular separation) between them lead to an improved calibration, as indicated by a lower Expected Calibration Error (ECE) (Fig.~\ref{fig:cosine_sim_hard_prompt}). Armed with this insight, we propose to impose orthogonalization constraints on the textual features by enforcing the angular distance between them. As such, this allows us to effectively utilize the feature space. Due to improved text feature separation, we observe superior calibration performance (Fig.~\ref{fig:radar_plot}). 
Our contributions are summarized as follows:

\begin{itemize}
\item We provide new insights underlying the suboptimal performance of an existing top-performing calibration method for the prompt testing-time tuning of VLMs. 

\item We propose a novel approach (named \texttt{\texttt{O-TPT}}) for calibrating test-time prompt tuning of VLMs by enforcing orthogonality constraints. This is accomplished by introducing orthogonal regularization on the textual features.  
\item We perform an extensive evaluation to validate our approach in various data sets and in different baselines. The results reveal that our \texttt{O-TPT} delivers consistent gains over the state-of-the-art methods in overall average calibration performance with several different baselines. Moreover, our \texttt{O-TPT} provides better calibration performance than the zero-shot CLIP, which reveals improved calibration compared to existing SOTA.

\end{itemize}


\section{Related Work}
\label{sec:related_work}

\noindent\textbf{Prompt tuning for vision-language models (VLMs):}
VLMs like CLIP \cite{radford2021learning} and ALIGN \cite{jia2021scaling} leveraged large datasets of image-text pairs to create shared embedding spaces that align images with their textual descriptions \cite{menonvisual}. 
This cross-modal alignment enables impressive zero-shot performance across various domains by applying hand-crafted prompt templates. As a result, models can predict without additional training on new tasks. 
The embeddings are optimized through contrastive learning by matching image-text pairs in a latent space, which enhances the model's ability to understand the relationship between visual and textual data. 
Since the design of handcrafted prompts requires domain-specific heuristics, they could be suboptimal for various newer domains. This led to the development of prompt-tuning methods that treat prompts as trainable embeddings.
A notable example is CoOP \cite{zhou2022learning}, which refined prompts in CLIP \cite{radford2021learning} using labeled samples, leading to better classification performance. 
Building upon this, CoCoOP \cite{zhou2022conditional} addressed the limitations of CoOP \cite{zhou2022learning} by enhancing generalization to out-of-distribution data, effectively improving the model's ability to new, unseen domains.
Recent advancements in prompt tuning, such as Test-Time Prompt Tuning (TPT) \cite{shu2022test}, have introduced algorithms to refine prompts on-the-fly using a single example during inference using entropy minimization. TPT has successfully enhanced model accuracy in zero-shot scenarios. However, it ends up worsening the calibration performance by making overconfident predictions.

\noindent\textbf{Calibration of deep neural networks:}
Calibration evaluates how closely a model's predicted confidence aligns with its accuracy. It is crucial for high-stakes applications that demand trustworthy uncertainty estimates, such as healthcare and autonomous systems \cite{ghahramani2015probabilistic}. 
Post hoc calibration techniques, such as Temperature Scaling (TS) \cite{guo2017calibration}, are employed to improve prediction reliability by scaling the model's logits using a temperature variable. This temperature variable is learned on a hold-out validation set.
However, such post-hoc methods often depend on the availability of labeled datasets that closely match the target data distribution \cite{liu2022devil}. Such datasets are rarely available in zero-shot and out-of-distribution contexts. 
To address this challenge, train-time calibration methods have been proposed, which typically employ auxiliary loss as regularizers with the task-specific loss during model training.
Some notable examples of train-time calibration methods for object classification \cite{liu2022devil,hebbalaguppe2022stitch, noh2023rankmixup} and object detection \cite{munir2022towards, munir2023bridging, munir2024cal} 
focused on reducing model overconfidence, leading to enhanced reliability in model predictions. However, these are supervised train-time calibration methods and are not directly applicable to scenarios where data is unlabeled, such as in the test-time prompt tuning of CLIP-like models. 



\noindent\textbf{Calibration \& VLMs:}
VLMs show good performance in zero-shot settings due to their extensive training on large-scale image-text pairs. 
However, when adapting these models to new domains or tasks, they often demonstrate poor calibration \textit{i.e.} the prediction confidence does not correspond to the actual likelihood of correctness. 
Methods based on test time prompting such as TPT \cite{shu2022test} can improve task-specific accuracy, but could increase model overconfidence by expanding the logit range.
To address these calibration challenges, \cite{murugesanrobust} introduced logit normalization techniques. 
Zero-shot logit normalization refines the logits in reference to (original) zero-shot settings, thereby maintaining the calibration of the model's confidence levels. 
Continuing in similar lines, \cite{murugesanrobust} proposed sample adaptive logit scaling, which normalizes logits per sample during inference. 
The work of \cite{yoon2024c} examined the dispersion of text features in VLMs and suggested that well-calibrated prompts tend to produce text embeddings with a broader dispersion between different classes \cite{yoon2024c}. 
By maximizing this dispersion, the calibration error can be reduced without compromising accuracy. To capture this characteristic, they proposed Average Text Feature Dispersion (ATFD) loss as a quantitative measure. 

However, we observe that this approach could be limited in establishing enough dispersion in challenging cases. This primarily happens due to underutilization of the space, and so the textual features can lie quite close to each other.
We develop a new approach that introduces orthogonalization constraints on textual features by enforcing the angular distance between them. This allows us to effectively disperse textual features, and hence better utilize the space, leading to consistently better calibration performance.





\begin{figure*}[!t]
    \centering
    \begin{floatrow}
        \ffigbox{%
            \includegraphics[width=\linewidth]{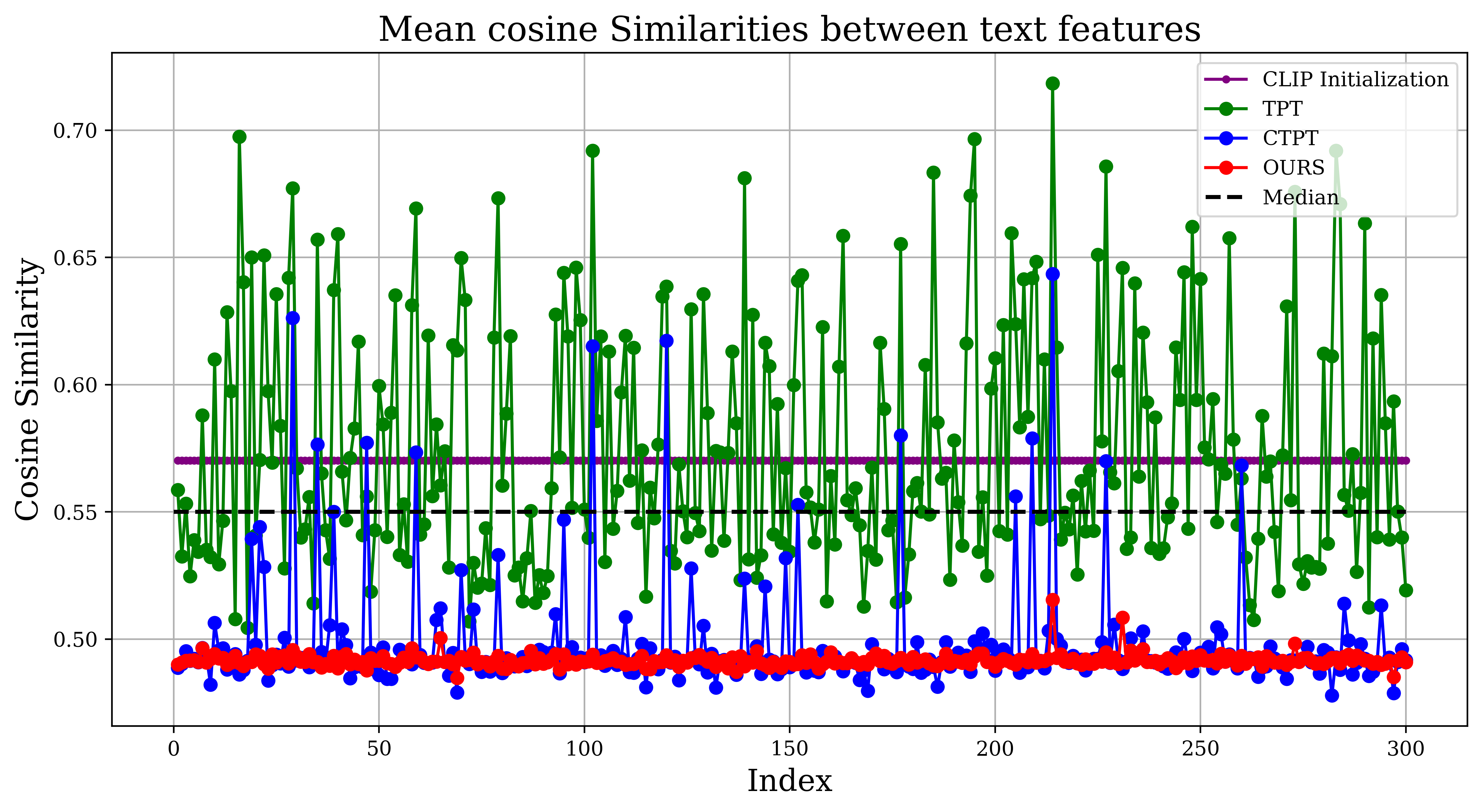}
        }{%
        \vspace{-2.5em}
            \caption{Mean cosine similarity changes comparison on a finegrained dataset \cite{nilsback2008automated} with CLIP B/16 backbone. Our orthogonal constraint offers consistent cosine similarity values among text features for all the data points.}
            \label{fig:mean_cos_sim}
        }
        
        \capbtabbox{%
            \small
            \begin{tabular}{l|l|ccc}
                \toprule
                Method & Metric & Group 1 & Group 2 & Overall \\
                \midrule
                \multirow{2}{*}{TPT} & Acc & 62.80 & 79.09 & 69.10 \\
                 & ECE & 15.97 & 9.45 & 13.45 \\
                \midrule
                \multirow{2}{*}{CTPT} & Acc & 63.66 & 79.51 & 69.79 \\
                 & ECE & 6.04 & 4.27 & 5.06 \\
                \midrule
                \multirow{2}{*}{\texttt{O-TPT} (Ours)} & \ccl Acc & \ccl  62.87 & \ccl  79.62 &  \ccl 70.07 \\
                 &  \ccl ECE &  \ccl \textbf{4.75}&  \ccl \textbf{3.80} &  \ccl \textbf{3.87} \\
                \bottomrule
            \end{tabular}
        }{%
            \caption{Comparison of Accuracy and Expected Calibration Error (ECE) across methods and categories based on the TPT\cite{shu2022test} text features cosine similarity.}
            \label{tab:calib_compare_above_init_below_init} \vspace{-1em}
        }
    \end{floatrow}
\end{figure*}

\section{Methodology}
\label{sec:Methodology}

\subsection{Preliminaries}
\label{subsec:Preliminaries}
\textbf{CLIP-based Zero-shot Classification:}  CLIP \cite{radford2021learning} is a vision-language model that learns a joint embedding space for images and texts by contrastive training on a vast dataset of image-text pairs. 
It consists of two encoders: a text encoder $\mathbf{f}_T$ and an image encoder $\mathbf{f}_I$, which map text and image inputs into a shared space.
In the zero-shot classification setting, CLIP classifies images by leveraging language prompts to represent each class. 
For each class $c_i$ in the set of classes $C = \{c_1, c_2, c_3, \dots, c_N\}$, we construct a textual prompt $\mathbf{t}_{c_i}$ using a hand-crafted template such as \emph{``a photo of a \{class\}''}, where ``\{class\}'' is class name.
Each text prompt $\mathbf{t}_{c_i}$ is fed into the text encoder to obtain a text feature vector: $\mathbf{e}_{c_i} = \mathbf{f}_T(\mathbf{t}_{c_i}), \text{for } i = 1, 2, \dots, N.$ This results in a set of text embeddings $\{\mathbf{e}_{c_1}, \mathbf{e}_{c_2}, \dots, \mathbf{e}_{c_N}\}$ that represent the classes in the shared embedding space. 
A test image $I$ is processed through the image encoder to produce an image embedding: $\mathbf{v} = \mathbf{f}_I(I).$
To determine how well the image matches each class, we compute the cosine similarity between the image embedding $\mathbf{v}$ and each text embedding $\mathbf{e}_{c_i}$: $s_i = \cos(\mathbf{v}, \mathbf{e}_{c_i})$.
Higher scores indicate greater similarity and alignment in the embedding space.
The similarity scores are converted into probabilities using the Softmax function with a temperature parameter $\tau$.
The temperature $\tau$ controls the smoothness of the Softmax distribution; a smaller $\tau$ sharpens it, increasing confidence in top predictions by amplifying score differences.
The predicted class $\hat{c}$ for the image $I$ is the one with the highest probability: $\hat{c} = \arg\max_{c_i} P(c_i|I).$ The corresponding predicted confidence is: $\hat{P} = \max_{c_i} P(c_i|I).$
This enables an efficient classification across a wide range of categories without the need for additional fine-tuning.

Transitioning from hand-crafted prompts, prompt tuning has been applied in CLIP \cite{Zhou_2022,zhou2022conditional,chen2023plotpromptlearningoptimal,Yao_2023_CVPR,radford2021learning} to optimize text prompts using labeled ImageNet samples, achieving robust cross-dataset generalization. Alternatively, TPT \cite{shu2022test} fine-tunes prompts at inference without labeled data, though it may amplify calibration error due to overconfidence \cite{pmlr-v70-guo17a}.

\noindent\textbf{Defining and Measuring Calibration Error:}
Calibration is crucial for a model because it ensures that its predicted probabilities correspond accurately to the actual likelihood of the outcomes. We can define this as: $\mathbb{P}(\hat{c}=c|\hat{P}=p)=p$.
Here, the input image $I$ and its corresponding ground truth label $c$ have an associated predicted confidence $p = \hat{P}$, and the predicted class is $\hat{c}$.
Calibration can be evaluated using the Expected Calibration Error (ECE) \cite{PakdamanNaeini_Cooper_Hauskrecht_2015}, which is computed as:
\begin{equation}
    \text{ECE} = \sum_{m=1}^M \frac{|A_m|}{N} \left| \text{acc}(A_m) - \text{conf}(A_m) \right|,
\end{equation}
where $M$ defines the number of bins used to divide the predictions, $A_m$ is the sample set with predicted confidence scores falling into the $m$-th bin. $|A_m|$ represents the number of samples in bin. $N$ is the total number of predictions. $\text{acc}(A_m)$ is the accuracy of the predictions and $\text{conf}(A_m)$ is the average predicted confidence for the samples in bin $A_m$.

\subsection{Orthogonal Constraint for Prompt Calibration}
\label{Orthogonality Constraint for Prompt Calibration}

\noindent\textbf{Why orthogonality?} Different prompts can yield text features with similar classification accuracy, but their calibration performance can differ significantly. 
It has been demonstrated, well-calibrated text features are often more dispersed concerning L2 distance, which is spread farther apart in embedding space  \cite{yoon2024c}.
This analysis overlooked an important factor: angular relationships among text features, which can significantly affect calibration.
We hypothesize that considering angular relationships among text features is important for understanding calibration of VLMs. Unlike existing methods (e.g., \cite{yoon2024c}) that rely solely on L2 distance for feature separation, our approach highlights significance of orthogonality among textual features.

To address this gap, we investigate the relationship between the cosine similarity of text features generated from various hand-crafted prompt styles. Our findings reveal a strong correlation: lower angular similarity between features corresponds to improved calibration. 
By encouraging greater angular separation (orthogonality), we ensure each feature points in a unique direction, sharpening class boundaries, and enhancing calibration robustness. 
As illustrated in Fig.~\ref{fig:cosine_sim_hard_prompt}, the probability density functions (PDFs) of cosine similarities across text feature pairs show that prompts yielding lower ECE are skewed toward lower cosine similarities. 
This suggests that promoting greater angular distinctiveness among text features can more effectively improve VLM calibration.



\noindent\textbf{Comparison between dispersion and orthogonality:} 
The ATFD loss in C-TPT \cite{yoon2024c} aims to disperse text features by moving their text features farther apart from their centroid. 
Still, it mainly affects the centroid’s position without ensuring adequate spacing between the individual features.
In zero-shot CLIP inference, normalized text features reside on the surface of the unit hypersphere\cite{shi2023towards}. In ATFD, this could potentially drive the centroid of the text features toward the origin of the hypersphere by merely adjusting the features on the surface.
%
However, this method may not effectively optimize the separation of features across the hypersphere's surface.
Towards achieving optimal calibration, we hypothesize that mere feature dispersion is insufficient and it is more effective to accomplish distinct angular separation between the features. 
To further validate our findings, we conducted an experiment on a fine-grained dataset \cite{nilsback2008automated}. For each data sample, we extract test-time prompt-tuned text features generated by TPT, C-TPT and our method (\texttt{\texttt{O-TPT}}), compute pairwise cosine similarities, and plot their mean, as shown in Fig.~\ref{fig:mean_cos_sim}. 
The results show that TPT, which lacks calibration-specific constraints, exhibits high fluctuations in cosine similarity, reflecting its inconsistent calibration performance.

In Table~\ref{tab:calib_compare_above_init_below_init}, we present the accuracy and ECE results for each method, dividing data into two groups based on cosine similarities of TPT text features relative to median cosine similarity. Group 1 includes points with cosine similarity values above median, while Group 2 includes those below the median. We then calculate the ECE and accuracy separately for each group, allowing a more fine-grained analysis of each method’s performance.  As hypothesized, points with higher cosine similarity tend to show elevated ECE, indicating poor calibration and suggesting these are more challenging points. 
In these challenging cases (Group 1), our method significantly outperforms TPT as well as C-TPT in terms of calibration performance, resulting in an overall lower ECE. Interestingly, C-TPT, which applies dispersion in the L2 space, also struggles to calibrate, showing higher cosine similarities in cases where TPT fails (these challenging points), as illustrated in Fig.~\ref{fig:ctpt_vs_ours_diag}. In contrast, our method’s orthogonalization constraint consistently produces text features with  much lower and more consistent cosine similarities compared to CLIP initialization, resulting in better calibration overall.
Our orthogonalization method enforces angular distance between feature pairs, fully utilizing the hyperspherical space (Fig.~\ref{fig:ctpt_vs_ours_diag}). 
As such, promoting orthogonality enhances feature separation, leading to distinct class boundaries and improved calibration.



\noindent\textbf{Orthogonalization constraint:} 
Motivated by the aforementioned analyses, we introduce an orthogonalization-based approach that systematically enhances calibration by focusing on the angular properties of the text feature matrix. 
For each class $c_i$, we have a text feature vector $\mathbf{e}_{c_i} \in \mathbb{R}^D$.
Let $\mathbf{E}$ be the text feature matrix containing text features of all classes, $\mathbf{E} \in \mathbb{R}^{C \times D}$. $E_{ij}$ represents the embedding value for $i$-th class in $j$-th dimension, where $i \in \{1, \dots, C\}$ indexes the classes and $j \in \{1, \dots, D\}$ represents the embedding dimensions. 
This matrix encapsulates the spatial distribution of text features between different classes. 
To enhance angular distances, we consider the matrix product $\mathbf{EE}^T$, which contains pairwise cosine similarities, a direct measure of the angular distance, between the text features. 
Ideally, we want $\mathbf{EE}^T$ to approximate the identity matrix $I_C$, indicating that all text features are orthogonal. 
To this end, we formulate a robust method by incorporating an orthogonalization constraint into the prompt tuning process, enhancing calibration performance of the test-time prompt tuning.
The objective formulation is defined as:
\begin{equation}
    \mathbf{t}^* = \arg \underset{\mathbf{t}}{\min}(L_{TPT}+ \lambda \lVert \mathbf{EE}^T - I_C \lVert_{2}^2)
\end{equation}
where $L_{TPT}$ is the TPT loss, $\lambda$ is a hyperparameter balancing the two terms.
This adjustment promotes a stable and uniformly distributed set of features across the hypersphere, directly addressing the ATFD loss's angular spacing limitations and ensuring effective utilization of the full feature space. 
By explicitly enforcing orthogonality, we systematically enhance the angular separation between text features, leading to improved calibration.

\section{Experiments}
\label{sec:Experiments}

\subsection{Experimental Setup}
\label{subsec:Experimental Setup}

\begin{table*}[h!]
\centering
\small
\setlength{\tabcolsep}{3pt} 
\renewcommand{\arraystretch}{1.2} 
\definecolor{lightgray}{gray}{0.9}
\scalebox{0.9}{
\begin{tabularx}{\textwidth}{l|c|*{12}{>{\centering\arraybackslash}X}} 
\toprule
\textbf{Method} & \textbf{Metric} & \rotatebox{90}{\textbf{INet}} & \rotatebox{90}{\textbf{DTD}} & \rotatebox{90}{\textbf{FLW}} & \rotatebox{90}{\textbf{Food}} & \rotatebox{90}{\textbf{SUN}} & \rotatebox{90}{\textbf{Air}} & \rotatebox{90}{\textbf{Pets}} & \rotatebox{90}{\textbf{Calt}} & \rotatebox{90}{\textbf{UCF}} & \rotatebox{90}{\textbf{SAT}} & \rotatebox{90}{\textbf{Car}} & \rotatebox{90}{\textbf{Avg}} \\
\midrule

\multirow{2}{*}{Zero Shot} 
& Acc. & 66.7  & 44.3 & 67.3 & 83.6 & 62.5 & 23.9 & 88.0 & 92.9 & 65.0 & 41.3 & 65.3 & 63.7 \\
& ECE & 2.12 & 8.50 & 3.00 & 2.39 & 2.53 & 5.11 &  4.37 & 5.50 & 3.59 & 13.89 & 4.25 & 4.43 \\
\midrule

\multirow{2}{*}{TPT} 
& Acc. & 69.0  & 46.7 & 69.0 & 84.7 & 64.5 & 23.4 & 87.1 & 93.8 & 67.3 & 42.4 & 66.3 & 65.0 \\
& ECE & 10.6 & 21.2 & 13.5 & 3.98 & 11.3 & 16.8 & 5.77 & 4.51 & 2.54 & 13.2 & 5.16 & 11.6 \\
\midrule

\multirow{2}{*}{C-TPT} 
& Acc & 68.5 & 46 & 69.8 & 83.7 & 64.8 & 24.85 & 88.2 & 93.63 & 65.7 & 43.2 & 65.8 & 64.57 \\
& ECE & 3.15 & 11.9 & 5.04 & 3.43 & 5.04 & 4.36 & 1.9 & 4.24 & 2.54 & 13.2 & 1.59 & 5.13 \\
\midrule



\multirow{2}{*}{Robust-adapt-SaLs-CTPT} & Acc.  & 68.04 & 45.51 & 69.43 & 83.18 & 64.38 & 23.94 & 88.12 & 93.63 & 65.32 &43.05 & 65.48 & 64.55 \\
 & ECE  & 2.63 & 14.56 & 2.74 & 1.26 & 3.56 & 6.21 & 3.16 & 3.78 & 6.96 &14.92 & 2.82 & 5.69 \\
\midrule

\multirow{2}{*}{Robust-adapt-Penalty-CTPT} 
& Acc. & 68.04 & 45.69 & 69.55 & 83.28 & 64.36 & 23.91 & 87.95 & 93.47 & 65.32 & 44.06 & 65.53 & 64.65 \\
& ECE & 2.63 & 13.9 & 5.27 & 3.35 & 4.87 & 4.43 & 1.63 & 4.56 & 2.29 & 7.08 & 1.25 & 4.66 \\
\midrule

\multirow{2}{*}{Robust-adapt-ZS-CTPT} 
& Acc & 68.01 & 45.63 & 69.55 & 83.25 & 64.41 & 23.88 & 88.03 & 93.31 & 65.24 & 42.64 & 65.45 & 64.51 \\
& ECE & 3.01 & 12.35 & 4.94 & 3.8 & 5.16 & 4.31 & 2.06 & 4.34 & 2.17 & 12.23 & 1.7 & 5.09 \\

\bottomrule

\multirow{2}{*}{\texttt{O-TPT} (Ours)}  & \ccl Acc. & \ccl 67.33  & \ccl 45.68 & \ccl 70.07 & \ccl 84.13 & \ccl 64.23 & \ccl 23.64 & \ccl 87.95 & \ccl 93.95 & \ccl 64.16 & \ccl 42.84 & \ccl 64.53 & \ccl 64.41 \\
& \ccl ECE & \ccl 1.96 & \ccl 7.88 & \ccl 3.87 & \ccl 1.46 & \ccl 4.93 & \ccl 3.68 & \ccl 1.9 & \ccl 3.8 & \ccl 2.34 & \ccl 12.98 & \ccl 1.78 & \ccl \textbf{4.23} \\
\midrule


\end{tabularx}}
\caption{Comparison of calibration performance with CLIP-ViTB/16 backbone. The overall best-performing result is in bold.}
\label{tab:resultsvitb16fine}
\end{table*}

\begin{table*}[h!]
\centering
\small
\setlength{\tabcolsep}{3pt} 
\renewcommand{\arraystretch}{1.2} 
\scalebox{0.9}{
\begin{tabularx}{\textwidth}{l|c|*{12}{>{\centering\arraybackslash}X}} 
\toprule
\textbf{Method} & \textbf{Metric} & \rotatebox{90}{\textbf{INet}} & \rotatebox{90}{\textbf{DTD}} & \rotatebox{90}{\textbf{FLW}} & \rotatebox{90}{\textbf{Food}} & \rotatebox{90}{\textbf{SUN}} & \rotatebox{90}{\textbf{Air}} & \rotatebox{90}{\textbf{Pets}} & \rotatebox{90}{\textbf{Calt}} & \rotatebox{90}{\textbf{UCF}} & \rotatebox{90}{\textbf{SAT}} & \rotatebox{90}{\textbf{Car}} & \rotatebox{90}{\textbf{Avg}} \\
\midrule

\multirow{2}{*}{Zero Shot} 
& Acc. & 58.1  & 40.0 & 61.0 & 74.0 & 58.6 & 15.6 & 83.8 & 85.8 & 58.4 & 23.7 & 55.7 & 55.9 \\
& ECE & 2.09 & 9.91 & 3.19 & 3.11 & 3.54 & 6.45 &  5.91 & 4.33 & 3.05 & 15.4 & 4.70 & 5.61 \\
\midrule

\multirow{2}{*}{TPT} 
& Acc. & 60.7  & 41.5 & 62.5 & 74.9 & 61.1 & 17.0 & 84.5 & 87.0 & 59.5 & 28.3 & 58.0 & 57.7 \\
& ECE & 11.4 & 25.7 & 13.4 & 5.25 & 9.24 & 16.1 & 3.65 & 5.04 & 12.4 & 22.5 & 3.76 & 11.7 \\
\midrule

\multirow{2}{*}{C-TPT} & Acc. &  60.2 & 42.2 & 65.2 & 74.7 & 61.0 & 17.0 & 84.1 & 86.9 & 59.7 & 27.8 & 56.5 & 57.75 \\
 & ECE & 3.01 & 19.8 & 4.14 & 1.86 & 2.93 & 10.7 & 2.77 & 2.07 & 3.83 & 15.1 & 1.94 & 6.19 \\
\midrule




\multirow{2}{*}{Robust-adapt-SaLs-CTPT} & Acc. & 60.02 & 41.37 & 65.0 & 74.71 & 60.46 & 16.83 & 83.73 & 86.53 & 59.56 & 27.54 & 56.27 & 57.20 \\
 & ECE &  2.21 & 9.2 & 2.29 & 1.45 & 3.32 & 5.6 & 3.6 & 4.38 & 4.2 & 10.46 & 2.53 & 6.82 \\
\midrule

\multirow{2}{*}{Robust-adapt-Penalty-CTPT} & Acc. & 60.06 & 41.55 & 64.88 & 74.69 & 60.47 & 17.01 & 83.84 & 86.45 & 59.42 & 26.64 & 56.4 & 57.40 \\
 & ECE &  5.93 & 21.18 & 4.03 & 1.83 & 2.81 & 10.85 & 3.3 & 2.7 & 3.93 & 11.97 & 1.99 & 6.41 \\
\midrule

\multirow{2}{*}{Robust-adapt-ZS-CTPT} & Acc. &  60.0 & 41.72 & 64.92 & 74.07 & 60.37 & 16.71 & 83.61 & 86.65 & 59.6 & 27.64 & 56.26 & 57.41 \\
 & ECE &  2.85 & 15.01 & 3.39 & 3.44 & 8.24 & 7.21 & 6.11 & 4.66 & 4.13 & 9.91 & 4.85 & 6.34 \\

 \midrule

\multirow{2}{*}{\texttt{O-TPT} (Ours)} & \ccl Acc. & \ccl 58.97 & \ccl 41.9 & \ccl 65.61 & \ccl 74.22 & \ccl 60.85 & \ccl 16.77 & \ccl 83.4 & \ccl 86.86 & \ccl 58.84 & \ccl 28.35 & \ccl 56.44 & \ccl 57.47 \\
 & \ccl ECE & \ccl 3.1 & \ccl 16.53 & \ccl 2.5 & \ccl 1.2 & \ccl 3.2 & \ccl 8.18 & \ccl 3.5 & \ccl 2.75 & \ccl 2.6 & \ccl 14.71 & \ccl 1.69 & \ccl \textbf{5.45} \\

 
\bottomrule
\end{tabularx}}
\caption{Comparison of calibration performance with CLIP-RN50 backbone. The overall best-performing result is in bold.}
\label{tab:resultsRN50fine}
\end{table*}

\begin{table}[h!]
\centering
\small
\setlength{\tabcolsep}{3.5pt} 
\scalebox{0.9}{
\begin{tabular}{l|l|ccccc} 
\toprule
\textbf{Method} & \textbf{Metric} & \textbf{I-A} & \textbf{I-V2} & \textbf{I-R} & \textbf{I-S} & \textbf{Avg} \\
\midrule
\multirow{2}{*}{CLIP-ViT-B/16} & Acc. & 47.8 & 60.8 & 74.0 & 46.1 & 57.2 \\
& ECE & 8.61 & 3.01 & 3.58 & 4.95 & 5.04 \\
\midrule
\multirow{2}{*}{TPT} & Acc. & 52.6 & 63.0 & 76.7 & 47.5 & 59.9 \\
& ECE & 16.4 & 11.1 & 4.36 & 16.1 & 12.0 \\
\midrule
\multirow{2}{*}{C-TPT} & Acc. & 51.6 & 62.7 & 76.0 & 47.9 & 59.6 \\
& ECE & 8.16 & 6.23 & 1.54 & 7.35 & 5.82 \\
\midrule
\multirow{2}{*}{\texttt{O-TPT} (Ours)} & \ccl  Acc. & \ccl  49.87 & \ccl  61.65 & \ccl  72.55 & \ccl  47.12 & \ccl  57.80 \\
& \ccl  ECE & \ccl  7.22 & \ccl  3.97 & \ccl  1.46 & \ccl  6.87 & \ccl  \textbf{4.88} \\
\midrule
\midrule

\multirow{2}{*}{CLIP-RN50} & Acc. & 21.7 & 51.4 & 56.0 & 33.3 & 40.6 \\
& ECE  & 21.3 & 3.33 & 2.07 & 3.15 & 7.46 \\
\midrule
\multirow{2}{*}{TPT} & Acc. & 25.2 & 54.6 & 58.9 & 35.1 & 43.5 \\
& ECE  & 31.0 & 13.1 & 9.18 & 13.7 & 16.7 \\
\midrule
\multirow{2}{*}{C-TPT} & Acc. & 23.4 & 54.7 & 58.0 & 35.1 & 42.8 \\
 & ECE  & 25.4 & 8.58 & 4.57 & 9.70 & 12.1 \\
\midrule

\multirow{2}{*}{\texttt{O-TPT} (Ours)} & \ccl  Acc. & \ccl  23.07 & \ccl  53.11 &  \ccl  54.47 & \ccl  33.98 & \ccl  41.16 \\
 & \ccl  ECE  & \ccl  24.56 & \ccl  3.87 & \ccl  4.47 & \ccl  5.85 & \ccl  \textbf{9.69} \\

 \bottomrule

\end{tabular}}
\caption{Calibration performance comparison with TPT (baseline) using CLIP-B/16 (top) and CLIP- RN50 (bottom) backbones in natural distribution shifts datasets.} 
\vspace{-0.5em}
\label{tab:resultsvitood}
\end{table}

\begin{table*}[h!]
\centering
\small
\setlength{\tabcolsep}{3pt} 
\renewcommand{\arraystretch}{1.2} 
\scalebox{0.9}{
\begin{tabularx}{\textwidth}{l|c|*{6}{>{\centering\arraybackslash}X}} 
\toprule
\textbf{Method} & \textbf{Metric} & \textbf{DTD} & \textbf{FLW} & \textbf{Food} & \textbf{Caltech} & \textbf{Car} & \textbf{Avg}\\
\midrule
\multirow{2}{*}{C-TPT} & Std. Acc & 0.12 & 0.16 & 0.16 & 0.22 & 0.2 & 0.17 \\
 & Std. ECE & 0.24 & 0.18 & 0.24 & 0.12 & 0.19 & 0.194 \\
 \midrule
\multirow{2}{*}{\texttt{O-TPT} (Ours)} & \ccl  Std. Acc &  \ccl 0.14 & \ccl  0.11 & \ccl  0.03 & \ccl  0.10 & \ccl  0.19 & \ccl  0.11 \\
 &  \ccl Std. ECE & \ccl  0.17 & \ccl  0.25 & \ccl  0.14 & \ccl  0.20 & \ccl  0.10 & \ccl  \textbf{0.177} \\
\bottomrule
\end{tabularx}}
\caption{Standard deviation across three different seed runs.} 
\label{tab:resultsstd} \vspace{-0.5em}
\end{table*}

\begin{table*}[h!]
\centering
\small
\setlength{\tabcolsep}{3pt} 
\renewcommand{\arraystretch}{1.2} 
\scalebox{0.9}{
\begin{tabularx}{\textwidth}{l|c|*{11}{>{\centering\arraybackslash}X}} 
\toprule
\textbf{Method} & \textbf{Metric} & \rotatebox{90}{\textbf{DTD}} & \rotatebox{90}{\textbf{FLW}} & \rotatebox{90}{\textbf{Food}} & \rotatebox{90}{\textbf{SUN}} & \rotatebox{90}{\textbf{Air}} & \rotatebox{90}{\textbf{Pets}} & \rotatebox{90}{\textbf{Calt}} & \rotatebox{90}{\textbf{UCF}} & \rotatebox{90}{\textbf{SAT}} & \rotatebox{90}{\textbf{Car}} & \rotatebox{90}{\textbf{Avg}} \\
\midrule

\multirow{2}{*}{TPT+CoOp} & Acc.  & 44.5 & 68.7 & 83.8 & 65.6 & 20.0 & 89.1 & 94.0 & 67.2 & 40.6 & 65.6 & 63.91 \\
 & ECE  & 34.8 & 19.9 & 9.66 & 20.8 & 29.6 & 7.40 & 3.65 & 19.9 & 31.3 & 6.63 & 18.36 \\
\midrule
\multirow{2}{*}{TPT+CoOp+C-TPT} & Acc.  & 45.0 & 69.0 & 83.7 & 65.1 & 19.2 & 89.3 & 93.9 & 66.6 & 40.7 & 63.1 & 63.56 \\
 & ECE  & 21.0 & 10.2 & 4.49 & 11.8 & 21.5 & 2.12 & 1.66 & 12.0 & 13.2 & 2.45 & 10.04 \\
\midrule
\multirow{2}{*}{TPT+CoOp+\texttt{O-TPT}} & \ccl  Acc.  & \ccl  45.45 & \ccl  68.57 & \ccl  83.55 & \ccl  64.01 & \ccl  18.69 & \ccl  89.07 & \ccl  93.71 & \ccl  65.64 & \ccl  40.17 & \ccl  64.12 &  \ccl 63.14 \\
& \ccl  ECE  & \ccl  16.02 & \ccl  6.81 & \ccl  3.59 & \ccl  7.23 & \ccl  16.82 & \ccl  1.92 & \ccl  0.92 & \ccl  9.16 & \ccl  13.76 &  \ccl 2.85 & \ccl  \textbf{7.91} \\
\bottomrule
\end{tabularx}}
\vspace{-0.5em}
\caption{Calibration performance when using CoOP as a baseline and CLIP-ViT-B/16 backbone.}
\label{tab:resultscoop}
\end{table*}

\begin{table*}[h!]
\centering
\small
\setlength{\tabcolsep}{3pt} 
\renewcommand{\arraystretch}{1.2} 
\scalebox{0.9}{
\begin{tabularx}{\textwidth}{l|c|*{11}{>{\centering\arraybackslash}X}} 
\toprule
\textbf{Method} & \textbf{Metric} & \rotatebox{90}{\textbf{DTD}} & \rotatebox{90}{\textbf{FLW}} & \rotatebox{90}{\textbf{Food}} & \rotatebox{90}{\textbf{SUN}} & \rotatebox{90}{\textbf{Air}} & \rotatebox{90}{\textbf{Pets}} & \rotatebox{90}{\textbf{Calt}} & \rotatebox{90}{\textbf{UCF}} & \rotatebox{90}{\textbf{SAT}} & \rotatebox{90}{\textbf{Car}} & \rotatebox{90}{\textbf{Avg}} \\
\midrule

\multirow{2}{*}{TPT+MAPLE} & Acc. & 50.05 & 70.72 & 85.01 & 64.87 & 24.36 & 87.78 & 94.42 & 66.48 & 47.32 & 66.5 & 65.75 \\
 & ECE  & 11.8 & 11.63 & 1.78 & 8.47 & 10.58 & 1.79 & 2.38 & 7.41 & 9.42 & 4.14 & 6.94 \\
\midrule

\multirow{2}{*}{TPT+MAPLE+\texttt{O-TPT}} & \ccl  Acc.  & \ccl  49.11 &  \ccl 71.53 & \ccl  84.35 & \ccl  63.49 & \ccl  24 & \ccl  89.97 & \ccl  92.29 & \ccl  65.82 & \ccl  44.58 & \ccl  65.38 & \ccl  65.05 \\
 & \ccl  ECE  & \ccl  4.9 & \ccl  4.35 & \ccl  1.49 & \ccl  2.78 & \ccl  6.41 & \ccl  3.97 & \ccl  3.49 & \ccl  2.22 & \ccl  7.92 & \ccl  3.61 & \ccl \textbf{ 4.11} \\
\bottomrule
\end{tabularx}}
\vspace{-0.5em}
\caption{Calibration performance when using Maple as a baseline and CLIP-ViT-B/16 backbone.} 
\label{tab:resultsMaple}
\end{table*}

\begin{table*}[h!]
\centering
\small
\setlength{\tabcolsep}{3pt} 
\renewcommand{\arraystretch}{1.2} 
\definecolor{lightgray}{gray}{0.9}
\scalebox{0.9}{
\begin{tabularx}{\textwidth}{l|c|*{12}{>{\centering\arraybackslash}X}} 
\toprule
\textbf{Method} & \textbf{Metric} & \rotatebox{90}{\textbf{INet}} & \rotatebox{90}{\textbf{DTD}} & \rotatebox{90}{\textbf{FLW}} & \rotatebox{90}{\textbf{Food}} & \rotatebox{90}{\textbf{SUN}} & \rotatebox{90}{\textbf{Air}} & \rotatebox{90}{\textbf{Pets}} & \rotatebox{90}{\textbf{Calt}} & \rotatebox{90}{\textbf{UCF}} & \rotatebox{90}{\textbf{SAT}} & \rotatebox{90}{\textbf{Car}} & \rotatebox{90}{\textbf{Avg}} \\
\midrule


\multirow{2}{*}{\textbf{\texttt{O-TPT}} without HouseHolder}
& Acc & 67.32 & 45.69 & 69.18 & 84.25 & 64.11 & 23.82 & 88.03 & 93.35 & 65.16 & 42.83 & 65.00 & 64.34 \\
& ECE & 1.96 & 8.61 & 3.94 & 1.66 & 5.28 & 3.59 & 1.93 & 4.21 & 2.19 & 13.89 & 1.71 & 4.45 \\

\bottomrule

\multirow{2}{*}{\texttt{O-TPT} (Ours)}  & \ccl Acc. & \ccl 67.33  & \ccl 45.68 & \ccl 70.07 & \ccl 84.13 & \ccl 64.23 & \ccl 23.64 & \ccl 87.95 & \ccl 93.95 & \ccl 64.16 & \ccl 42.84 & \ccl 64.53 & \ccl 64.41 \\
& \ccl ECE & \ccl 1.96 & \ccl 7.88 & \ccl 3.87 & \ccl 1.46 & \ccl 4.93 & \ccl 3.68 & \ccl 1.9 & \ccl 3.8 & \ccl 2.34 & \ccl 12.98 & \ccl 1.78 & \ccl \textbf{4.23} \\
\midrule


\end{tabularx}}
\caption{Calibration performance of our \texttt{O-TPT} with and without HouseHolder transformation.} 
\vspace{-0.75em}
\label{tab:results_without_HT}
\end{table*}

\begin{table}[h!]
\centering
\small
\setlength{\tabcolsep}{1pt} 
\renewcommand{\arraystretch}{1.2} 
\scalebox{0.9}{
\begin{tabularx}{\textwidth}{l*{11}{>{\centering\arraybackslash}X}} 
\toprule
\textbf{Method}  & \rotatebox{0}{\textbf{DTD}} & \rotatebox{0}{\textbf{FLW}} & \rotatebox{0}{\textbf{Calt}} & \rotatebox{0}{\textbf{Avg}} \\
\midrule
\multirow{1}{*}{Zero-shot} 
   & 1.34 & 0.59 & 0.25 & 0.73  \\
\multirow{1}{*}{TPT} 
   & 1.42 & 0.50 & 0.16 & 0.69  \\
\multirow{1}{*}{C-TPT} 
   & 1.31 & 0.52 & 0.22 & 0.68 \\
\multirow{1}{*}{\texttt{O-TPT}} 
 &  \ccl 1.24 & \ccl  0.53 &  \ccl 0.17 & \ccl \textbf{ 0.65}  \\
\bottomrule
\end{tabularx}}
\caption{Static Calibration Error (SCE)  ($10^{-2}$) performance comparison with CLIP-ViT-B/16 backbone.} 
\label{tab:resultsSCE}
\end{table}

\noindent\textbf{Datasets:} We conduct our experiments using the ImageNet \cite{deng2009imagenet} and Caltech101 \cite{FeiFei2004LearningGV} datasets  as well as different fine-grained classification datasets across various domains. For texture classification, we use DTD \cite{cimpoi2014describing}. The FLW \cite{nilsback2008automated} dataset comprises of flower categories. For food classification, we use Food101 \cite{bossard14}. SUN37 \cite{xiao2016sun} dataset offers scene categorization, and the UCF\cite{soomro2012ucf101} dataset is used for human action recognition. Vehicle classification datasets include StanfordCars \cite{maji13fine-grained} and Aircraft \cite{maji2013fine}, while OxfordPets \cite{parkhi2012cats} covers pet animal categories. Additionally, EuroSAT \cite{helber2018introducing} contains satellite imagery for environmental categorization. To evaluate out-of-distribution (OOD) performance, we use the ImageNet-A \cite{hendrycks2019nae}, ImageNet-V2, ImageNet-R \cite{hendrycks2020many}, and ImageNet-S\cite{wang2019learning} datasets.

\noindent\textbf{Implementation details:}
We utilize the CLIP-RN50 and CLIP ViT-B/16 architectures. In all experimental settings, we leverage TPT \cite{shu2022test} as our baseline.
To optimize the prompt, we use a single-step update with the AdamW optimizer \cite{loshchilov2019decoupledweightdecayregularization}, with a learning rate of 0.005. We also perform a linear algebra technique, HouseHolder transform \cite{KAUFMAN1987221} on top of $\mathbf{EE}^T$ matrix to further enhance optimization at one test-time step. We use a batch size of 64 across all experiments.  All experiments were performed on an NVIDIA RTX A6000 with 48GB of memory. We initialize the prompt embeddings as hard prompts, following the settings in C-TPT \cite{yoon2024c}. The remaining settings follow the configuration in \cite{shu2022test}.  We fix the $\lambda$ as 18 across all experiments unless otherwise specified. 

\subsection{Results}
\label{subsec:Results}

We evaluate the calibration performance of our method (\texttt{O-TPT}) applied to TPT with CLIP-B/16 and CLIP-RN50 backbones (Tables~\ref{tab:resultsvitb16fine} \& \ref{tab:resultsRN50fine}). \texttt{O-TPT} which applies orthogonal constraints on text features significantly improves Expected Calibration Error (ECE) compared to both C-TPT \cite{yoon2024c} and Robust Adapt \cite{murugesanrobust} (in combination with C-TPT). Using the CLIP-B/16 backbone, our method achieves an average ECE of \textbf{4.21}, outperforming C-TPT at 5.13 and Robust Adapt’s best result of 4.66. When applied to the CLIP-RN50 backbone, our approach reduces ECE to \textbf{5.45}, a substantial improvement over the 6.19 ECE achieved by C-TPT. Notably, our method also surpasses the zero-shot calibration performance showing lower ECE on both backbones, which is a feat unmatched by any other approach, while maintaining the high accuracy levels of TPT.

\noindent\textbf{Results on natural distribution shifts:}
Table \ref{tab:resultsvitood} presents results on natural distribution shift datasets, where all experimental configurations remain the same as those detailed in the implementation, with the exception of $\lambda$, which is set to 2 for these datasets. Across the ImageNet variant datasets, our \texttt{O-TPT} method consistently achieves notable reductions in ECE on both the CLIP-RN50 and CLIP-ViT-B/16 backbones. On the CLIP-ViT-B/16 backbone, \texttt{O-TPT} achieves an average ECE of \textbf{4.88}, a substantial improvement over C-TPT and TPT, which average 5.82 and 12.0, respectively. Similarly, on the CLIP-RN50 backbone, \texttt{O-TPT} achieves an average ECE of \textbf{9.69}, outperforming C-TPT and TPT, which reach ECE values of 12.1 and 16.7, respectively. These results highlight the effectiveness of our approach in handling natural distribution shifts.


\subsection{Ablation Studies}
\label{subsec: Ablation Studies}



\begin{figure}[h]
    \centering
    \includegraphics[width=\linewidth]{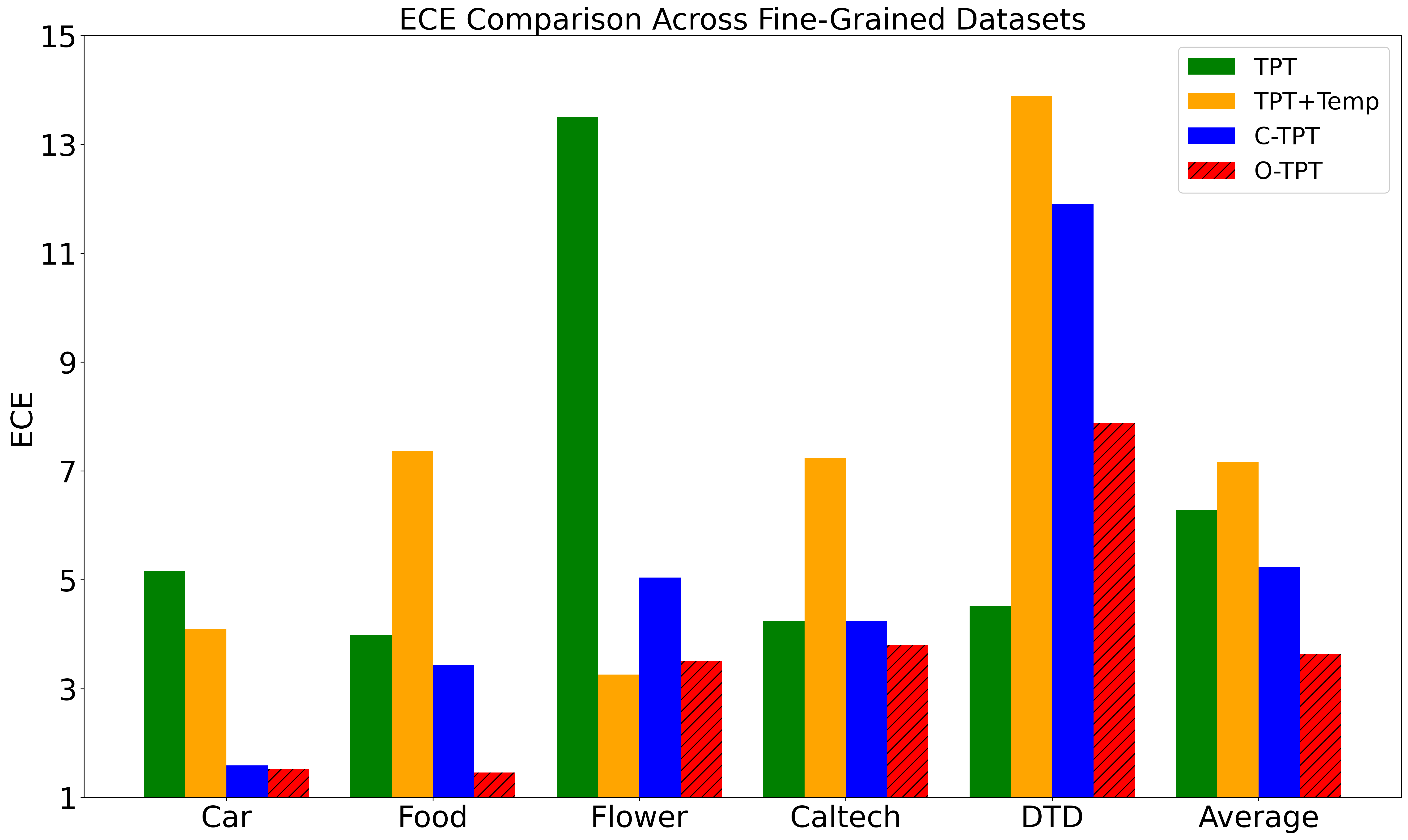}
    \caption{Calibration performance comparison with post-hoc method across fine-grained datasets.}
    \label{fig:comparision} \vspace{-1em}
\end{figure}

\noindent\textbf{Comparison with post-hoc method:} Fig. \ref{fig:comparision} illustrates the impact of ECE across fine-grained datasets using calibration techniques applied to TPT \cite{shu2022test}. We compare TPT, C-TPT \cite{yoon2024c}, TPT with temperature scaling \cite{pmlr-v70-guo17a} (a classical post-processing technique adjusting pre-softmax logits with a temperature value trained on a separate validation set), and our proposed method, \texttt{O-TPT}. As shown in Fig. \ref{fig:comparision}, \texttt{O-TPT} consistently achieves superior calibration, demonstrating its effectiveness in enhancing model reliability across datasets.

\begin{figure}[h]
    \centering
    \begin{subfigure}{0.32\linewidth}
        \centering
        \includegraphics[width=\linewidth]{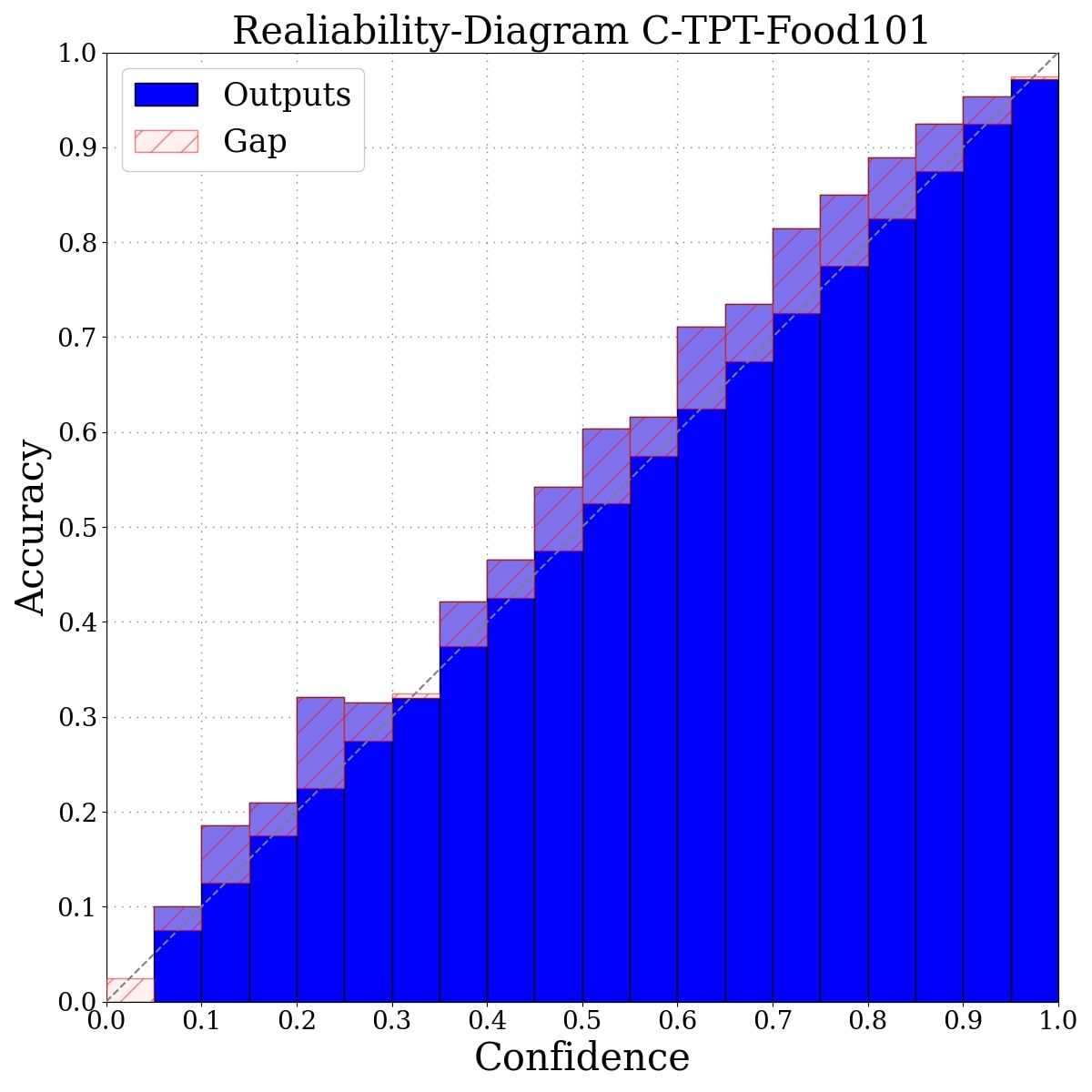}
        \caption{C-TPT: Food}
        \label{ctptfood}
    \end{subfigure}%
    \hfill
    \begin{subfigure}{0.32\linewidth}
        \centering
        \includegraphics[width=\linewidth]{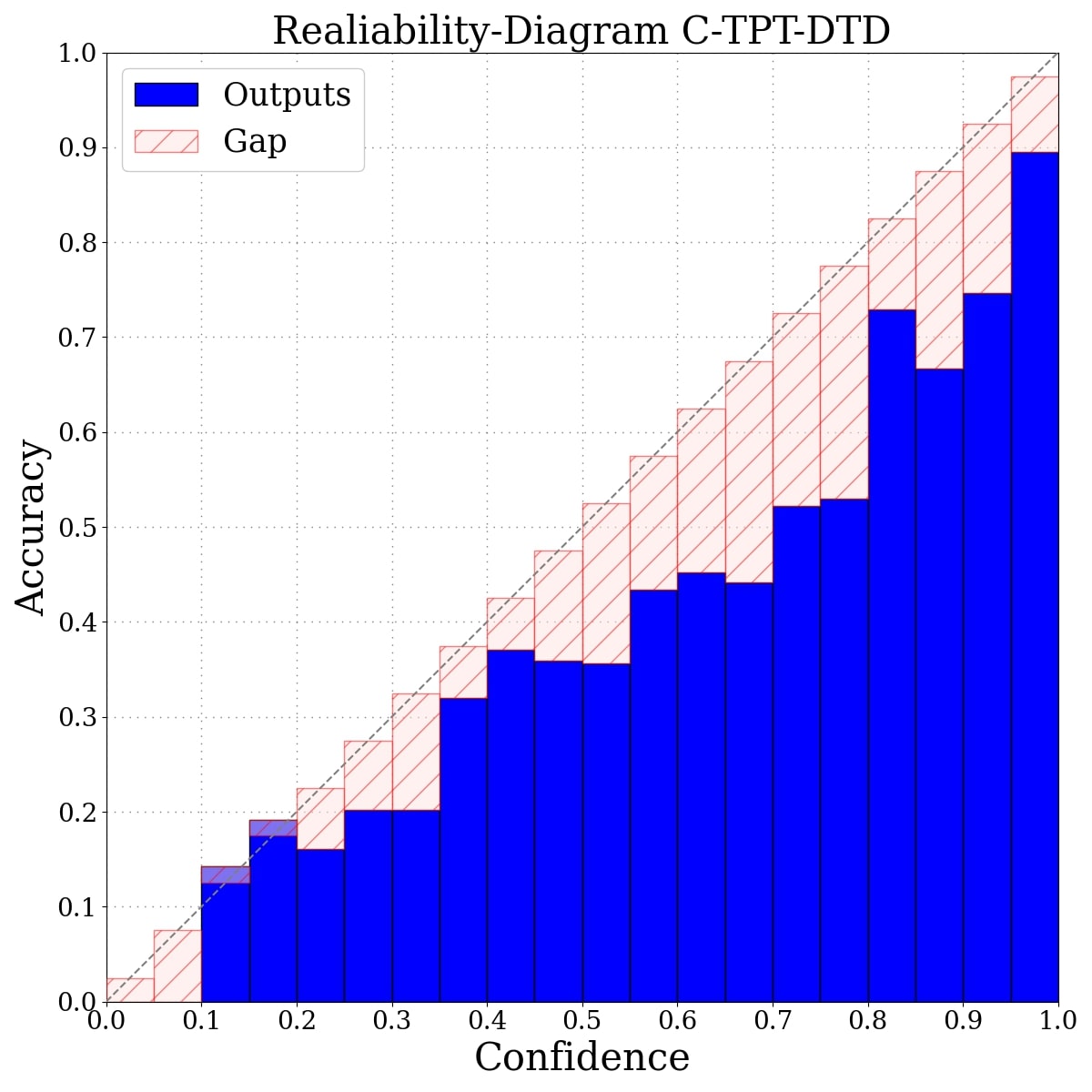}
        \caption{C-TPT: DTD}
        \label{ctptdtd}
    \end{subfigure}%
    \hfill
    \begin{subfigure}{0.32\linewidth}
        \centering
        \includegraphics[width=\linewidth]{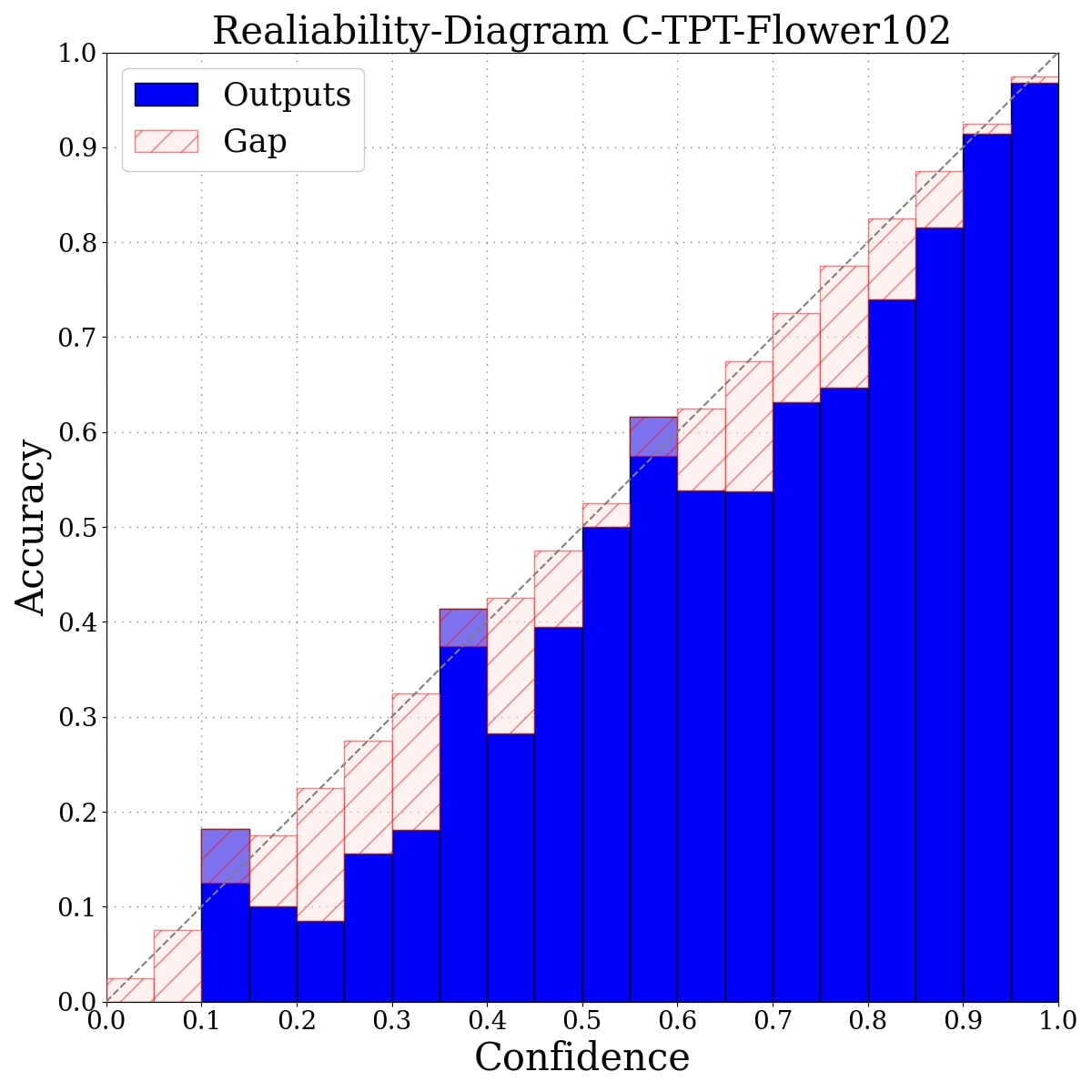}
        \caption{C-TPT: Flower}
        \label{ctptflower}
    \end{subfigure}%
    \hfill
    \begin{subfigure}{0.32\linewidth}
        \centering
        \includegraphics[width=\linewidth]{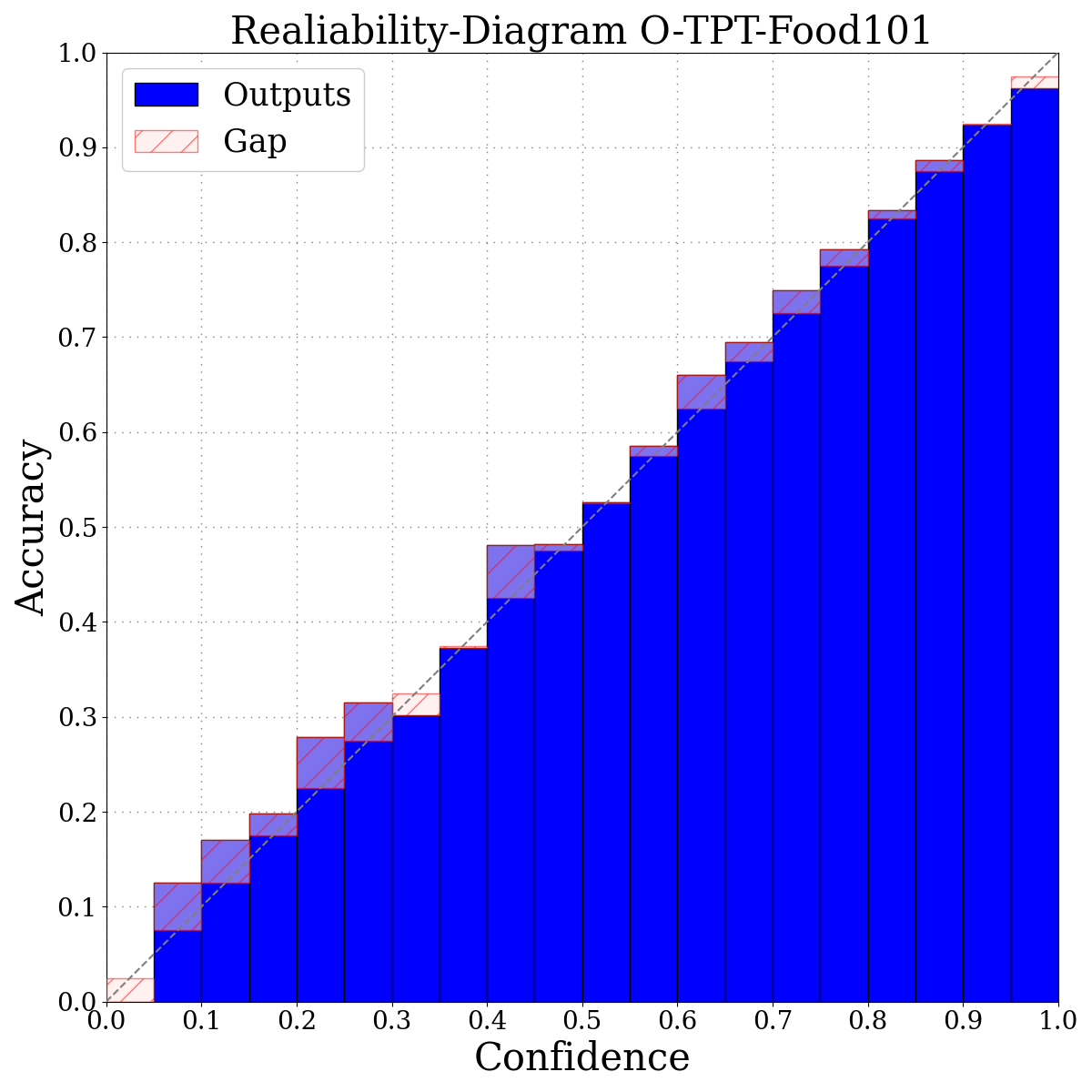}
        \caption{\texttt{O-TPT}:Food}
        \label{htfood}
    \end{subfigure}%
    \hfill
    \begin{subfigure}{0.32\linewidth}
        \centering
        \includegraphics[width=\linewidth]{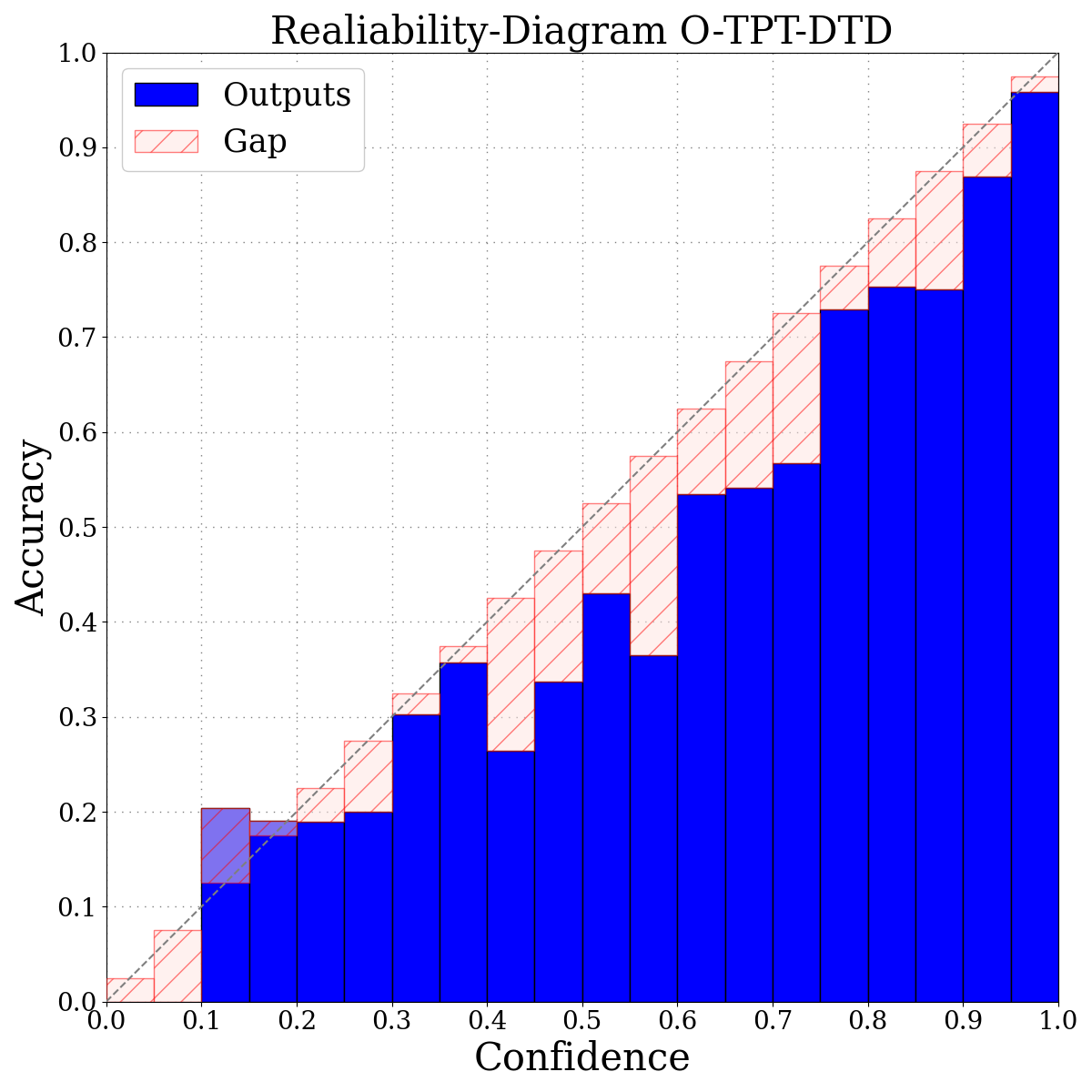}
        \caption{\texttt{O-TPT}:DTD}
        \label{htdtd}
    \end{subfigure}%
    \hfill
    \begin{subfigure}{0.32\linewidth}
        \centering
        \includegraphics[width=\linewidth]{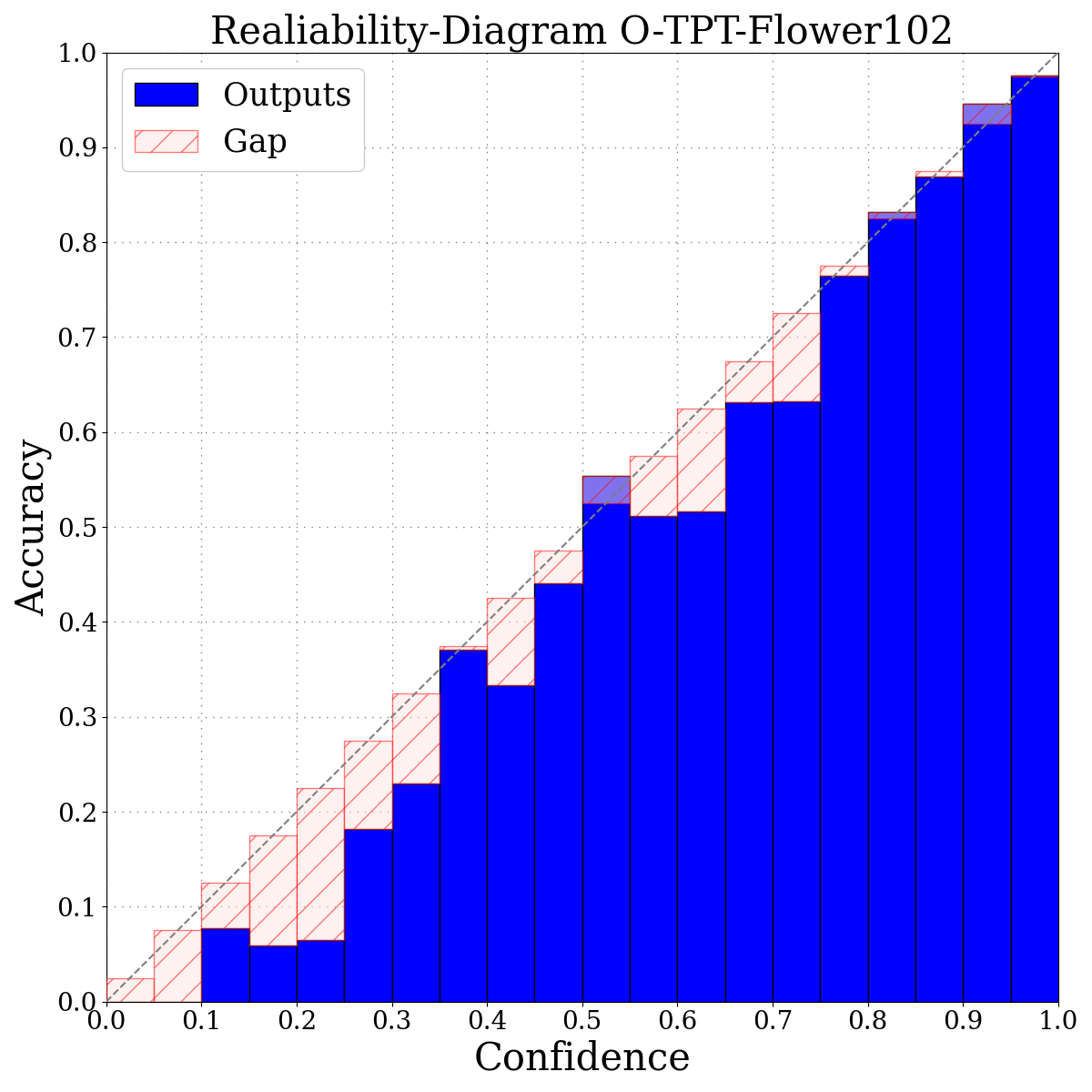}
        \caption{\texttt{O-TPT}:Flower}
        \label{htflower}
    \end{subfigure}
    \caption{Reliability diagrams.}
    \label{fig:ctpt_reliability_diagram} \vspace{-0.5em}
\end{figure}

\noindent\textbf{Addressing Under and Over-Confidence:} As shown in Tables \ref{tab:resultsvitb16fine} and \ref{tab:resultsRN50fine}, our proposed method outperforms most fine-grained datasets on both ResNet-50 and ViT-B/16 backbones. However, these tables do not indicate whether our approach addresses overconfidence or underconfidence. Fig.\ref{fig:ctpt_reliability_diagram}  displays reliability diagrams for ViT-B/16 on the Food, Flower, and DTD datasets. In Fig.\ref{ctptfood}, C-TPT \cite{yoon2024c} shows underconfidence for the Food dataset, which is rectified with our \texttt{O-TPT} calibration method, as shown in Fig.\ref{htfood}. Additionally, our method better addresses overconfidence issues, as illustrated in Fig.\ref{htdtd} and \ref{htflower}, when compared to Fig.\ref{ctptdtd} and \ref{ctptflower}, respectively.

\noindent\textbf{Standard deviation across different seeds run:} Table \ref{tab:resultsstd} shows the mean and standard deviation for accuracy and ECE across five datasets, each with three different seed runs. Our method reports a lower mean standard deviation in accuracy, providing more stable performance regardless of the randomness in the prompt initialization. Additionally, our method has a lower mean standard deviation in ECE compared to C-TPT, indicating greater consistency in calibration. This robustness in both accuracy and calibration makes our approach particularly valuable for applications where stable performance across multiple runs is essential.

\noindent\textbf{Other Baselines:} Beyond tuning prompt parameters at inference time, we applied supervised-trained prompt embedding parameters to evaluate their calibration effectiveness with our \texttt{O-TPT} method during test-time prompt tuning. Specifically, we used the officially published checkpoints of CoOp \cite{zhou2022learning} and MAPLE \cite{khattak2023maplemultimodalpromptlearning}. For MAPLE, we employed the Base-to-Novel checkpoints. As shown in Table \ref{tab:resultscoop}, for CoOp across 10 fine-grained datasets, our method demonstrates significant improvement on 8 datasets, reducing the overall average ECE to \textbf{7.91}. This represents a substantial improvement over C-TPT + CoOp and TPT + CoOp, which yield overall ECE values of 10.4 and 18.36, respectively. Similarly, when evaluating our method \texttt{O-TPT} with MAPLE under test-time prompt tuning, we observe notable calibration improvements across most datasets compared to using TPT + MAPLE alone. In Table \ref{tab:resultsMaple}, our approach achieves an overall ECE reduction to \textbf{4.11}, compared to 6.94 for TPT + MAPLE. Overall, integrating \texttt{O-TPT} with supervised-trained prompt embedding parameters during test-time prompt tuning maintains accuracy and significantly improves calibration performance. 

\noindent\textbf{Impact of HouseHolder transformation:} Table \ref{tab:results_without_HT} presents the impact of applying Householder transformation. 
This transformation further improves our performance. 

\noindent\textbf{SCE performance comparison:} In addition to ECE, we report calibration performance with Static Calibration Error (SCE) \cite{Nixon_2019_CVPR_Workshops}, a class-wise extension of ECE. Table \ref{tab:resultsSCE} compares the calibration performance of our method against other approaches. Across the three datasets, \texttt{O-TPT} consistently outperforms the alternatives, achieving an average SCE reduction of up to \textbf{0.65}, which marks a substantial improvement over Zero-shot, TPT, and C-TPT, with average SCE values of 0.73, 0.69, and 0.68, respectively. See supplementary for SCE results on additional datasets.

\section{Conclusion}
\label{sec:Conclusion}
We propose a new approach to calibrate test-time prompt tuning in vision-language models. Our preliminary analyses reveal that the higher angular distances between textual features while prompt learning are correlated to lower calibration error. We note that, achieving orthogonality between textual features is more effective than obtaining dispersion through L2 distance-based objectives. Therefore, we propose orthogonal regularization on textual features during test-time prompt tuning which is named as \texttt{\texttt{O-TPT}}. Our method consistently records state-of-the-art performance with different backbones and baselines.  

{
    \small
    \bibliographystyle{ieeenat_fullname}
    \bibliography{main}
}
\clearpage

\setcounter{page}{1}
\maketitlesupplementary
\renewcommand{\thesection}{A\arabic{section}}
\setcounter{section}{0}


\noindent In this supplementary material, we provide the following:

\begin{enumerate}
    \item We reveal the relation between calibration performance and angular distances (sec.~\ref{relation_eces_vs_calibration})
    \item We compare the SCE performance (sec.~\ref{sec:sce_compare})
    \item Reliabilty plots comparison with C-TPT \cite{yoon2024c}. (sec.~\ref{sec:reliabilit_plots})
    \item Calibration performance with different hard prompt styles (sec.~\ref{sec:diffpromp})
    \item Calibration with a Combination of C-TPT
and O-TPT (sec.~\ref{sec:c-tpt+o-tpt})
     \item  O-TPT results on Medical Prompt tunning
methods (sec.~\ref{sec:med+o-tpt})
    \item  Pareto Front analysis with varying $\lambda$ (sec.~\ref{sec:parato-analyse})  
    
\end{enumerate}

\section{Relation Between Calibration Performance and Angular Distances}
\label{relation_eces_vs_calibration}
\begin{figure}[!h]
    \centering
    \includegraphics[width=\linewidth]{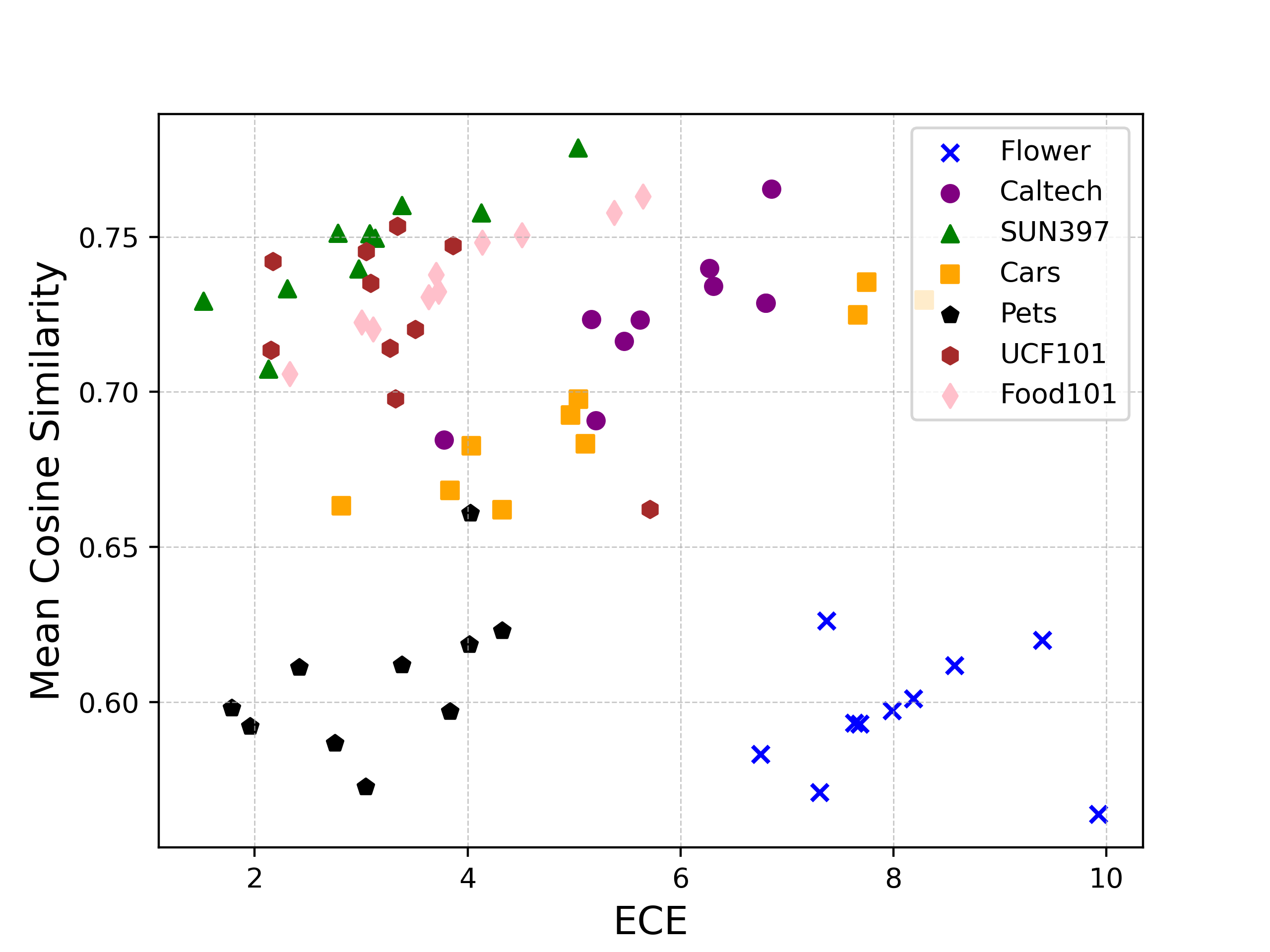}
    \caption{Relation of ECE with cosine similarities (of textual features) on CLIP-B/16 backbone.}
    \label{fig:ece vs cos sims}
\end{figure}

To further validate our motivation, we conduct an experiment using 80 different hard prompt styles \cite{radford2021learning} to evaluate their corresponding Expected Calibration Error (ECE) performance and the mean cosine similarity of text features. This evaluation is performed on zero-shot inference using the CLIP-B/16 backbone. Figure~\ref{fig:ece vs cos sims} illustrates the results for prompts that yield higher accuracies across seven diverse datasets: Flower, Caltech101, SUN397, Cars, Pets, UCF101, and Food101. Specifically, we focus on the top 10 prompt styles that provide the highest accuracies.

The results reveal a clear trend between mean cosine similarity (an angular distance measure) and ECE (calibration performance). A lower mean cosine similarity correlates with a reduced ECE, indicating that greater angular distancing among text features promotes better calibration. This suggests that prompts with text features exhibiting greater angular distances between their representations lead to improved calibration outcomes.

\section{SCE performance comparison}
\label{sec:sce_compare}
Tables \ref{tab:resultsscevit} and \ref{tab:resultsscern} present the Static Calibration Error (SCE) results across 10 datasets using CLIP-B/16 and CLIP RN-50 backbones. Our method, \texttt{O-TPT}, outperforms C-TPT on both backbones, achieving an overall average SCE values of \textbf{1.07} for CLIP-B/16 and \textbf{1.24} for CLIP RN-50, demonstrating improved calibration performance.

\begin{figure}[h]
    \centering
    \begin{subfigure}[t]{0.24\linewidth}
        \centering
        \includegraphics[width=\linewidth]{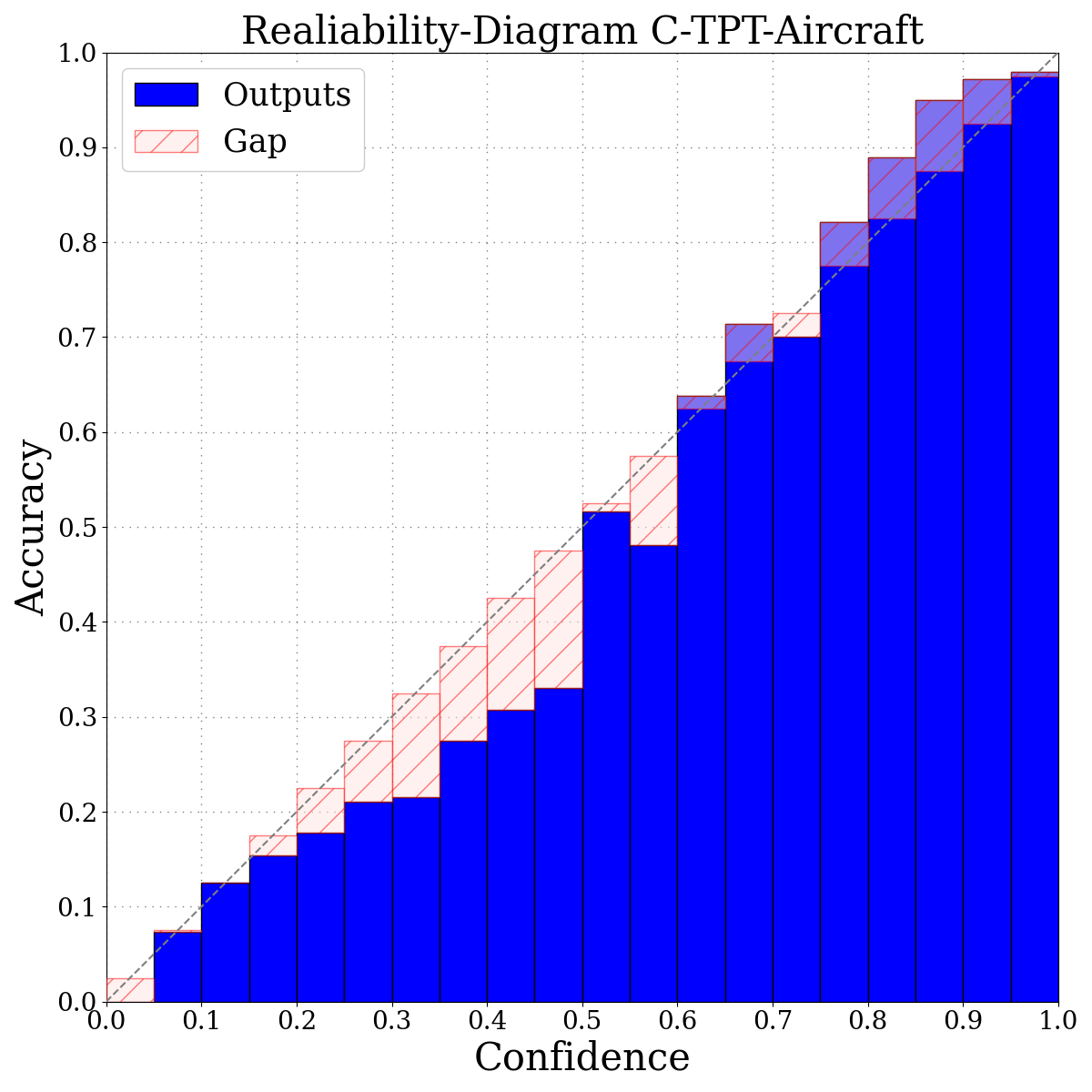}
        \caption{C-TPT:Air}
        \label{ctptAircraft}
    \end{subfigure}%
    \begin{subfigure}[t]{0.24\linewidth}
        \centering
        \includegraphics[width=\linewidth]{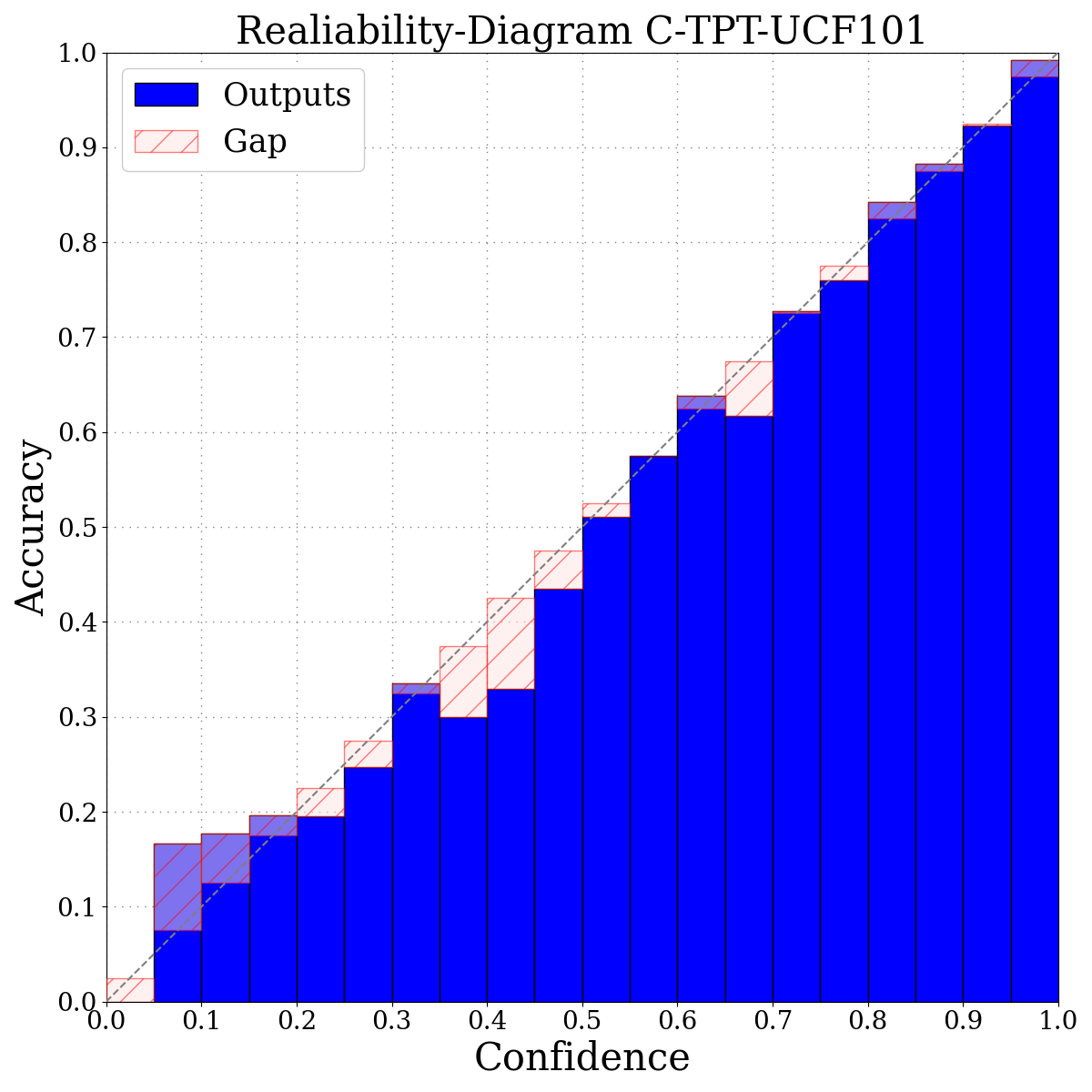}
        \caption{C-TPT:UCF}
        \label{ctptucf101}
    \end{subfigure}%
    \begin{subfigure}[t]{0.24\linewidth}
        \centering
        \includegraphics[width=\linewidth]{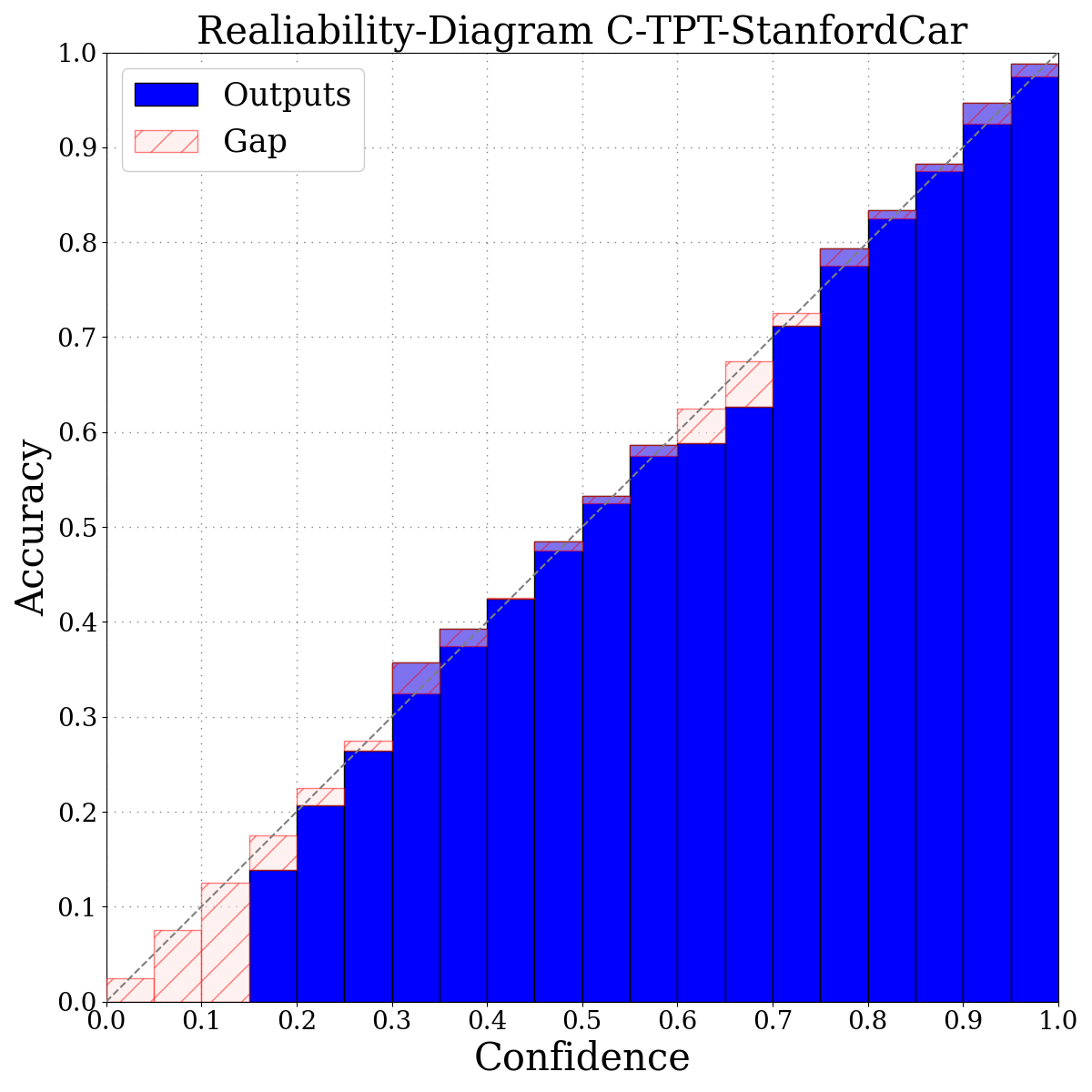}
        \caption{C-TPT:Car}
        \label{ctpteurosat}
    \end{subfigure}%
    \begin{subfigure}[t]{0.24\linewidth}
        \centering
        \includegraphics[width=\linewidth]{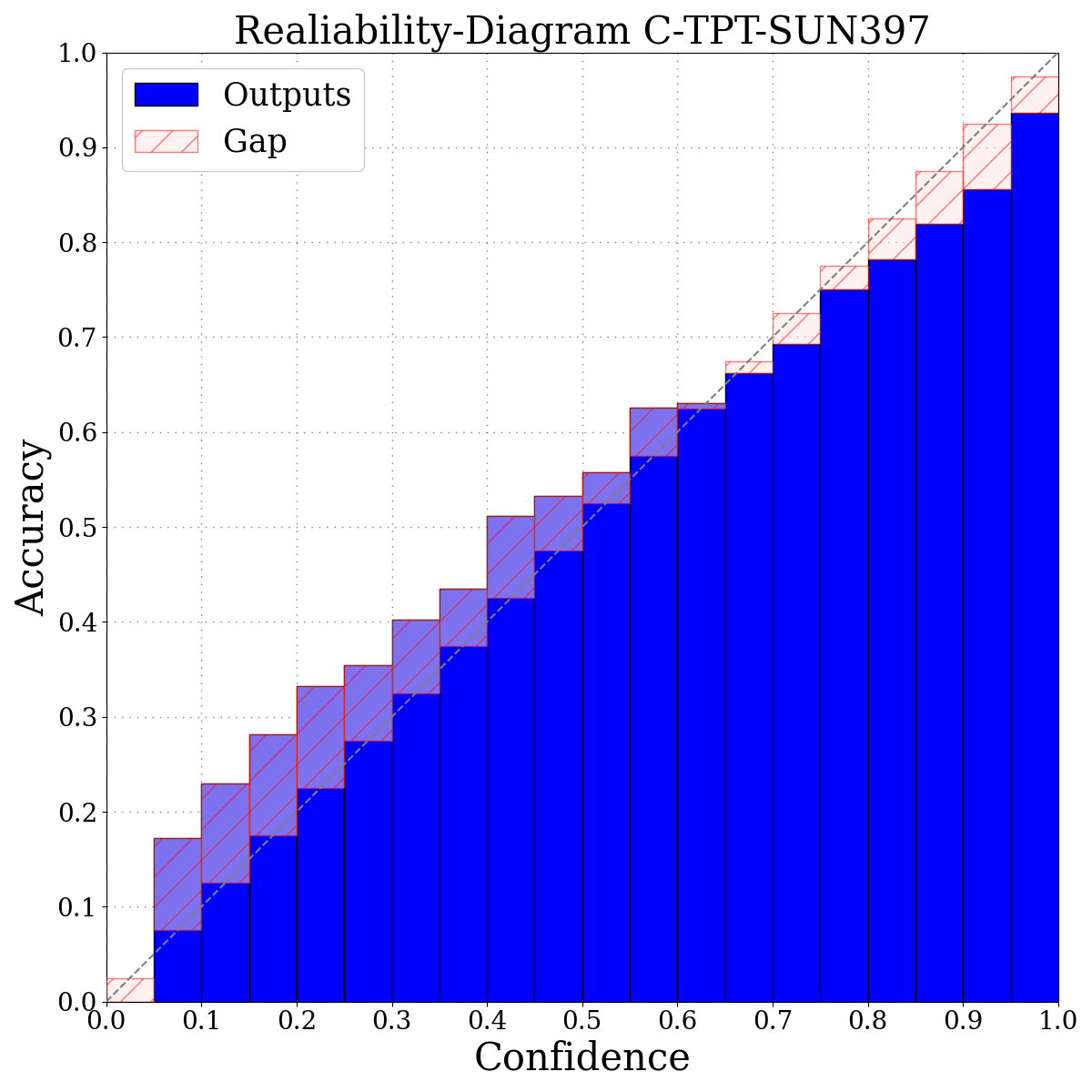}
        \caption{C-TPT:SUN}
        \label{ctptsun}
    \end{subfigure}%
    
    \vspace{0.5em} 

    \begin{subfigure}[t]{0.24\linewidth}
        \centering
        \includegraphics[width=\linewidth]{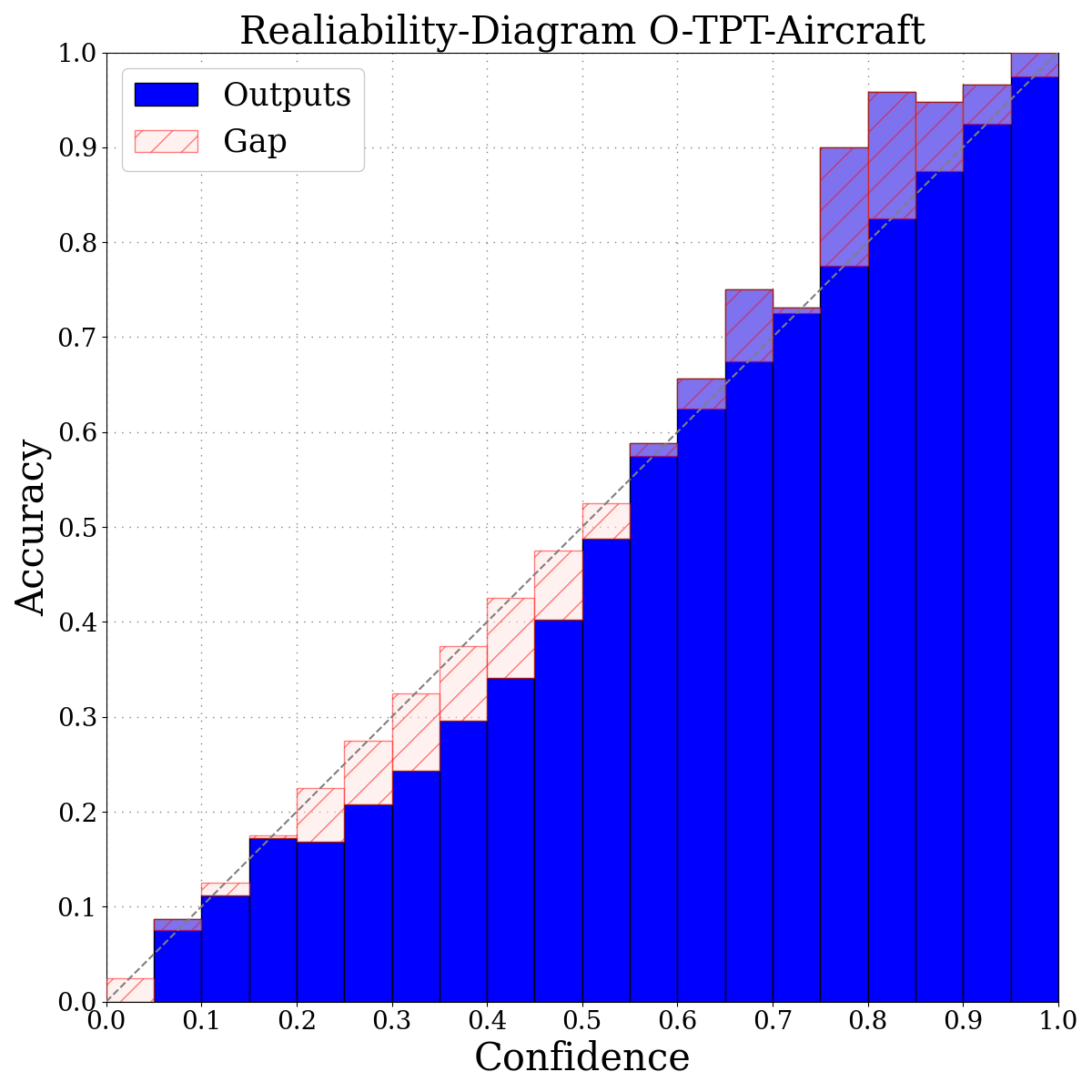}
        \caption{\texttt{O-TPT}:Air}
        \label{htaircraft}
    \end{subfigure}%
    \begin{subfigure}[t]{0.24\linewidth}
        \centering
        \includegraphics[width=\linewidth]{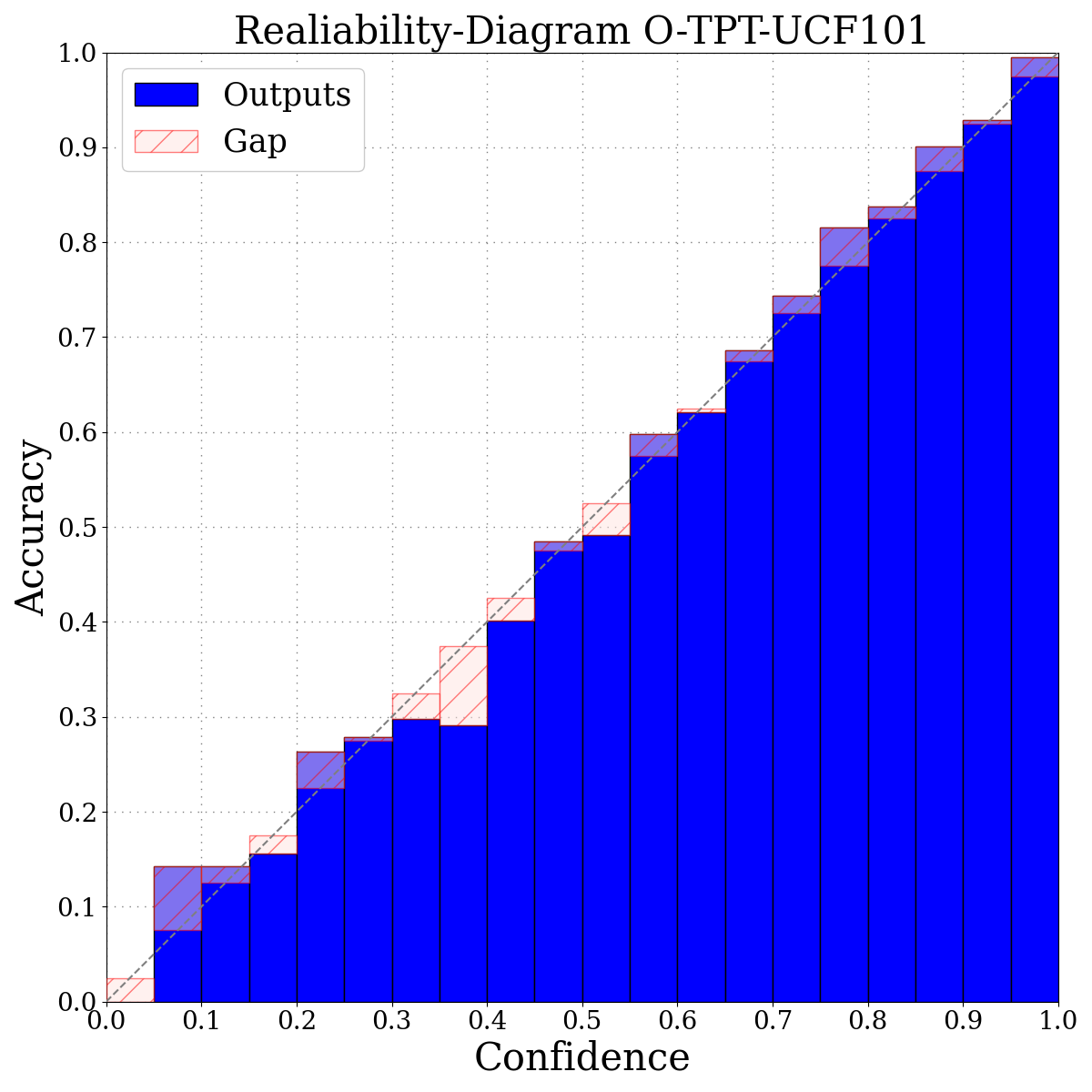}
        \caption{\texttt{O-TPT}:UCF}
        \label{htucf101}
    \end{subfigure}%
    \begin{subfigure}[t]{0.24\linewidth}
        \centering
        \includegraphics[width=\linewidth]{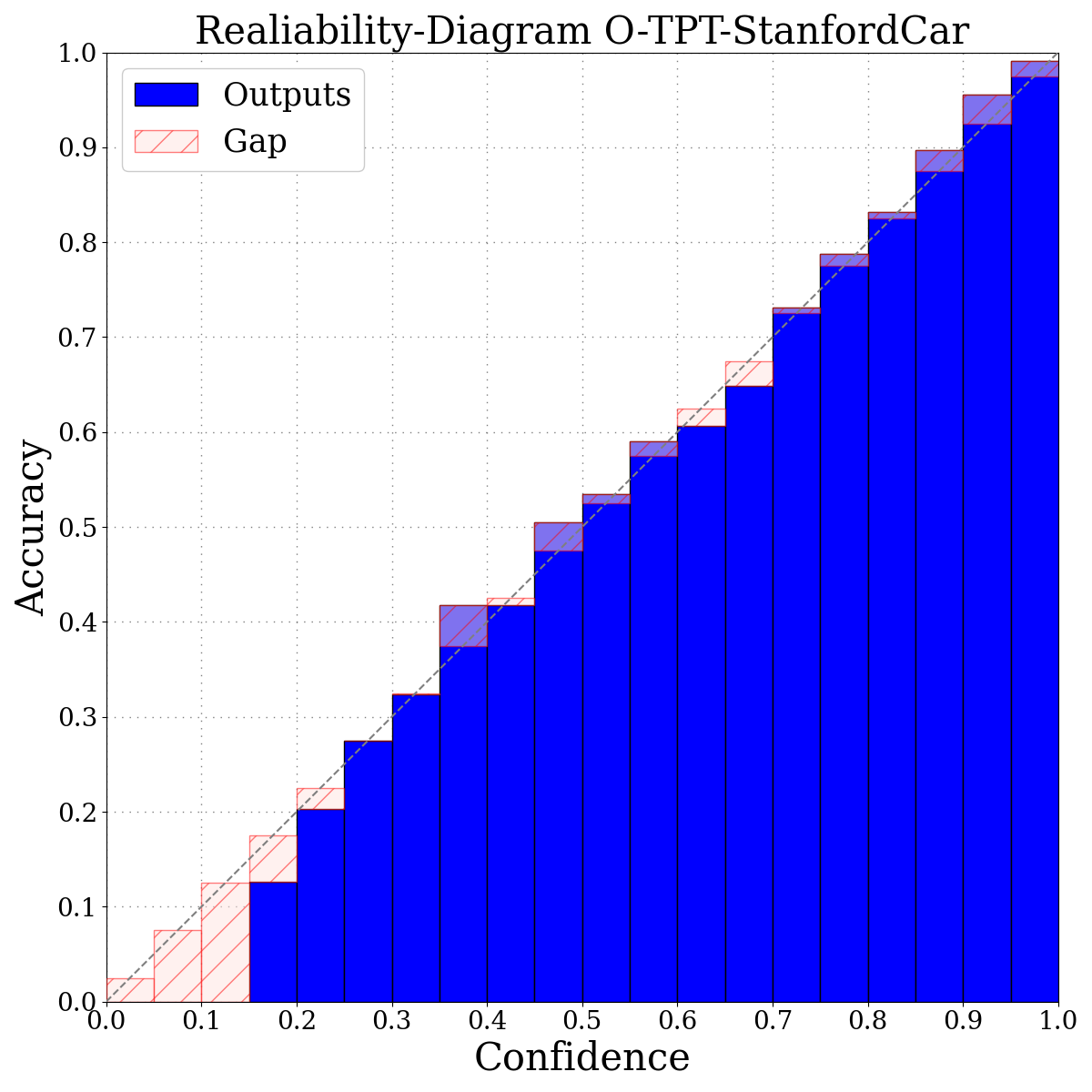}
        \caption{\texttt{O-TPT}:Car}
        \label{hteurosat}
    \end{subfigure}%
    \begin{subfigure}[t]{0.24\linewidth}
        \centering
        \includegraphics[width=\linewidth]{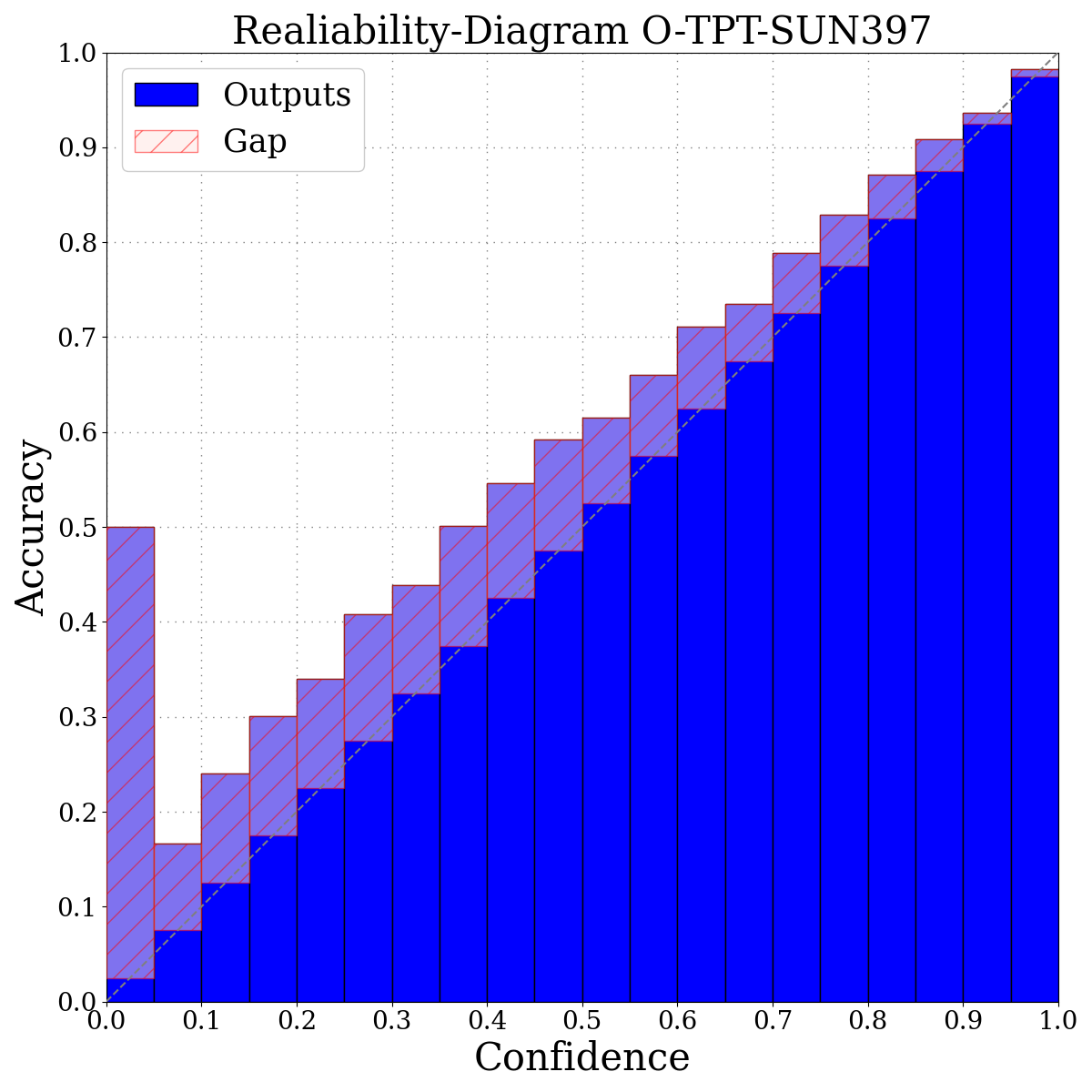}
        \caption{\texttt{O-TPT}:SUN}
        \label{htsun}
    \end{subfigure}%
    
    \caption{Reliability diagrams for CLIP-B/16.}
    \label{fig:vit_reliability_diagram}
    \vspace{-0.5em}
\end{figure}

\begin{table*}[h!]
\centering
\small
\setlength{\tabcolsep}{3pt} 
\renewcommand{\arraystretch}{1.2} 
\definecolor{lightgray}{gray}{0.9}
\scalebox{0.9}{
\begin{tabularx}{\textwidth}{l|c|*{12}{>{\centering\arraybackslash}X}} 
\toprule
\textbf{Method} & \textbf{Metric} &  \rotatebox{90}{\textbf{DTD}} & \rotatebox{90}{\textbf{FLW}} & \rotatebox{90}{\textbf{Food}} & \rotatebox{90}{\textbf{SUN}} & \rotatebox{90}{\textbf{Air}} & \rotatebox{90}{\textbf{Pets}} & \rotatebox{90}{\textbf{Calt}} & \rotatebox{90}{\textbf{UCF}} & \rotatebox{90}{\textbf{SAT}} & \rotatebox{90}{\textbf{Car}} & \rotatebox{90}{\textbf{Avg}} \\
\midrule

\multirow{1}{*}{Zero Shot} 
& SCE  & 1.33 & 0.59 & 0.20 & 0.12 & 0.52 & 0.68  & 0.25 & 0.52 & 6.18 & 0.23 & 1.06 \\
\midrule

\multirow{1}{*}{TPT} 

& SCE & 1.44 & 0.51 & 0.17 & 0.15 & 0.58 & 0.60 & 0.16 & 0.57 & 7.07 & 0.25 & 1.15 \\
\midrule

\multirow{1}{*}{C-TPT} 

& SCE & 1.31 & 0.52 & 0.22 & 0.14 & 0.56 & 0.58 & 0.22 & 0.52 & 6.81 & 0.22 & 1.11 \\

\bottomrule

\multirow{1}{*}{\texttt{O-TPT} (Ours)}  
& \ccl SCE & \ccl 1.24 & \ccl 0.53 & \ccl 0.19 & \ccl 0.12 & \ccl 0.56 & \ccl 0.57 & \ccl 0.17 & \ccl 0.51 & \ccl 6.58 & \ccl 1.07 & \ccl \textbf{1.07} \\
\midrule


\end{tabularx}}
\caption{Static Calibration Error (SCE)  ($10^{-2}$) performance comparison with CLIP-B/16 backbone.}
\label{tab:resultsscevit}
\end{table*}

\begin{table*}[h!]
\centering
\small
\setlength{\tabcolsep}{3pt} 
\renewcommand{\arraystretch}{1.2} 
\definecolor{lightgray}{gray}{0.9}
\scalebox{0.9}{
\begin{tabularx}{\textwidth}{l|c|*{12}{>{\centering\arraybackslash}X}} 
\toprule
\textbf{Method} & \textbf{Metric} &  \rotatebox{90}{\textbf{DTD}} & \rotatebox{90}{\textbf{FLW}} & \rotatebox{90}{\textbf{Food}} & \rotatebox{90}{\textbf{SUN}} & \rotatebox{90}{\textbf{Air}} & \rotatebox{90}{\textbf{Pets}} & \rotatebox{90}{\textbf{Calt}} & \rotatebox{90}{\textbf{UCF}} & \rotatebox{90}{\textbf{SAT}} & \rotatebox{90}{\textbf{Car}} & \rotatebox{90}{\textbf{Avg}} \\
\midrule

\multirow{1}{*}{Zero Shot} 
& SCE  & 1.31 & 0.66 & 0.29 & 0.12 & 0.54 &  0.73 & 0.35 & 0.54 & 7.39 & 0.23 & 1.22 \\
\midrule

\multirow{1}{*}{TPT} 

& SCE & 1.52 & 0.63 & 0.25 & 0.11 & 0.60 & 0.54 & 0.38 & 0.51 & 8.23 & 0.24 & 1.30 \\
\midrule

\multirow{1}{*}{C-TPT} 

& SCE & 1.43 & 0.62 & 0.26 & 0.11 & 0.53 & 0.67 & 0.32 & 0.51 & 8.07 & 0.23 & 1.27 \\

\bottomrule

\multirow{1}{*}{\texttt{O-TPT} (Ours)}  
& \ccl SCE & \ccl 1.34 & \ccl 0.60 & \ccl 0.27 & \ccl 0.12 & \ccl 0.51 & \ccl 0.69 & \ccl 0.3 & \ccl 0.5 & \ccl 7.85 & \ccl 0.22 & \ccl \textbf{1.24} \\
\midrule


\end{tabularx}}
\caption{Static Calibration Error (SCE)  ($10^{-2}$) performance comparison with CLIP-RN-50 backbone.}
\label{tab:resultsscern}
\end{table*}

\begin{table*}[h!]
\centering
\small
\setlength{\tabcolsep}{3pt} 
\renewcommand{\arraystretch}{1.2} 
\definecolor{lightgray}{gray}{0.9}
\scalebox{0.9}{
\begin{tabularx}{\textwidth}{l|c|*{12}{>{\centering\arraybackslash}X}} 
\toprule
\textbf{Method} & \textbf{Metric}  & \rotatebox{90}{\textbf{DTD}} & \rotatebox{90}{\textbf{FLW}} & \rotatebox{90}{\textbf{Food}} & \rotatebox{90}{\textbf{SUN}} & \rotatebox{90}{\textbf{Air}} & \rotatebox{90}{\textbf{Pets}} & \rotatebox{90}{\textbf{Calt}} & \rotatebox{90}{\textbf{UCF}} & \rotatebox{90}{\textbf{SAT}} & \rotatebox{90}{\textbf{Car}} & \rotatebox{90}{\textbf{Avg}} \\
\midrule

\multirow{2}{*}{Zero Shot} 
& Acc. & 38.4 & 64.5 & 81.4 & 62.4 & 22.7 & 86.2 & 88.1 & 67.6 & 34.6 & 66.5 & 61.24 \\
& ECE & 7.43 & 4.59 & 1.10 & 6.11 & 2.83 & 7.43 & 14.1 & 2.65 & 14.1 & 4.59 & 7.01 \\
\midrule

\multirow{2}{*}{TPT} 
& Acc. & 45.5 & 67.9 & 84.9 & 65.9 & 24.5 & 87.4 & 91.5 & 66.4 & 43.3 & 67.2 & 64.45 \\
& ECE & 20.0 & 14.6 & 5.74 & 13.3 & 19.2 & 6.34 & 3.11 & 14.1 & 18.2 & 6.36 & 12.09 \\
\midrule

\multirow{2}{*}{C-TPT} 
& Acc & 46.3 & 69.6 & 84.1 & 65.5 & 24.7 & 88.8 & 91.7 & 67.0 & 43.0 & 66.9 & 64.76 \\
& ECE & 18.0 & 10.6 & 2.43 & 10.7 & 10.5 & 1.59 & 1.89 & 7.42 & 8.73 & 1.64 & 7.35 \\

\bottomrule

\multirow{2}{*}{\texttt{O-TPT} (Ours)}  & \ccl Acc.   & \ccl 44.62 & \ccl 68.29 & \ccl 84.82 & \ccl 63.05 & \ccl 23.16 & \ccl 88.28 & \ccl 91.48 & \ccl 64.74 & \ccl 44.81 & \ccl 66.02 & \ccl 63.92 \\
& \ccl ECE  & \ccl 12.85 & \ccl 4.67 & \ccl 1.85 & \ccl 2.67 & \ccl 6.37 & \ccl 3.59 & \ccl 3.0 & \ccl 4.08 & \ccl 8.33 & \ccl 2.71 & \ccl \textbf{5.01} \\
\midrule


\end{tabularx}}
\caption{Comparison of calibration performance with CLIP-B/16 backbone with the prompt of  `a photo of the cool \{class\}'}
\label{tab:vitaphotoofthecool}
\end{table*}

\section{Reliability Plots}
\label{sec:reliabilit_plots}

\begin{figure}[!h]
    \centering
    \begin{subfigure}[t]{0.24\linewidth}
        \centering
        \includegraphics[width=\linewidth]{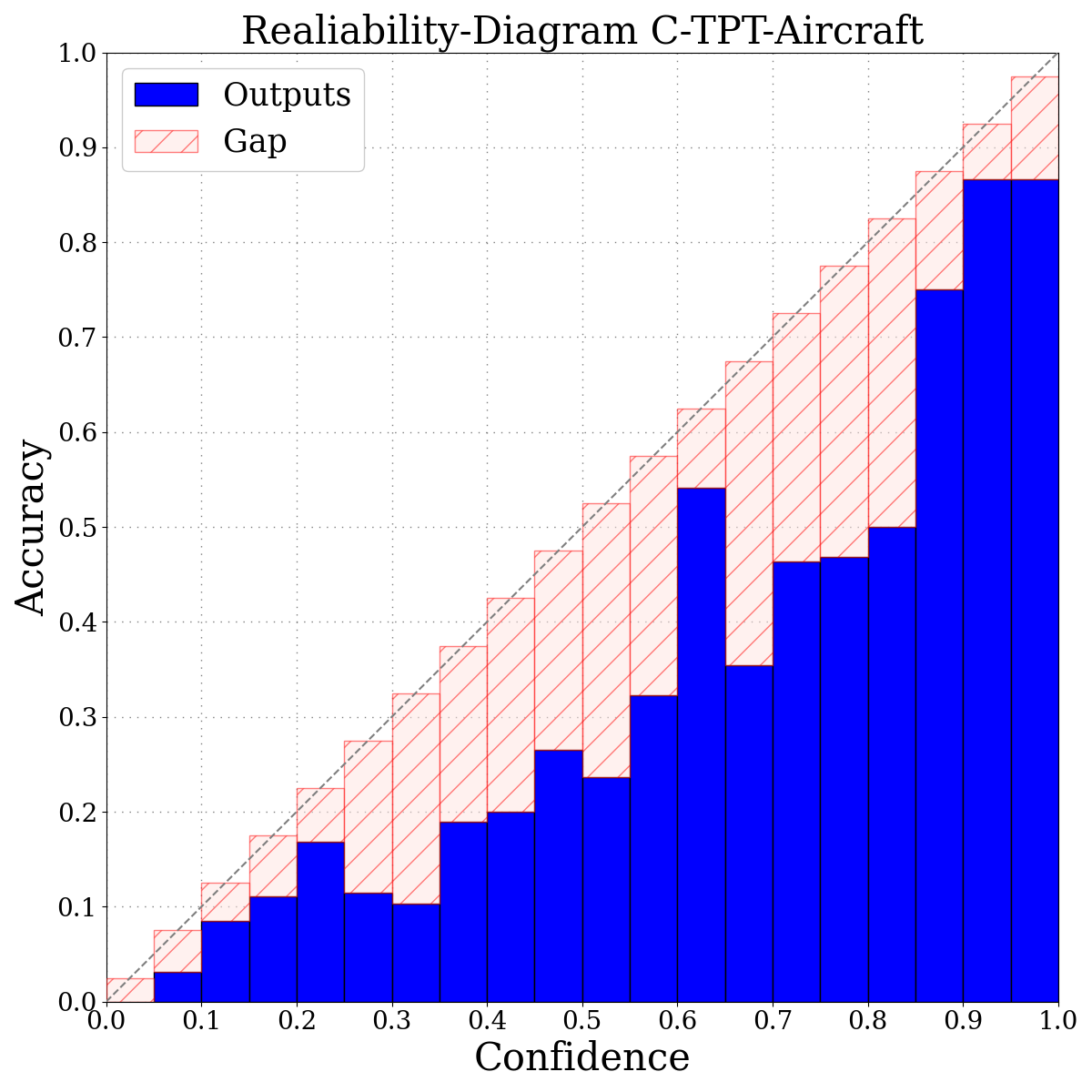}
        \caption{C-TPT:Air}
        \label{rnctptAircraft}
    \end{subfigure}%
    \begin{subfigure}[t]{0.24\linewidth}
        \centering
        \includegraphics[width=\linewidth]{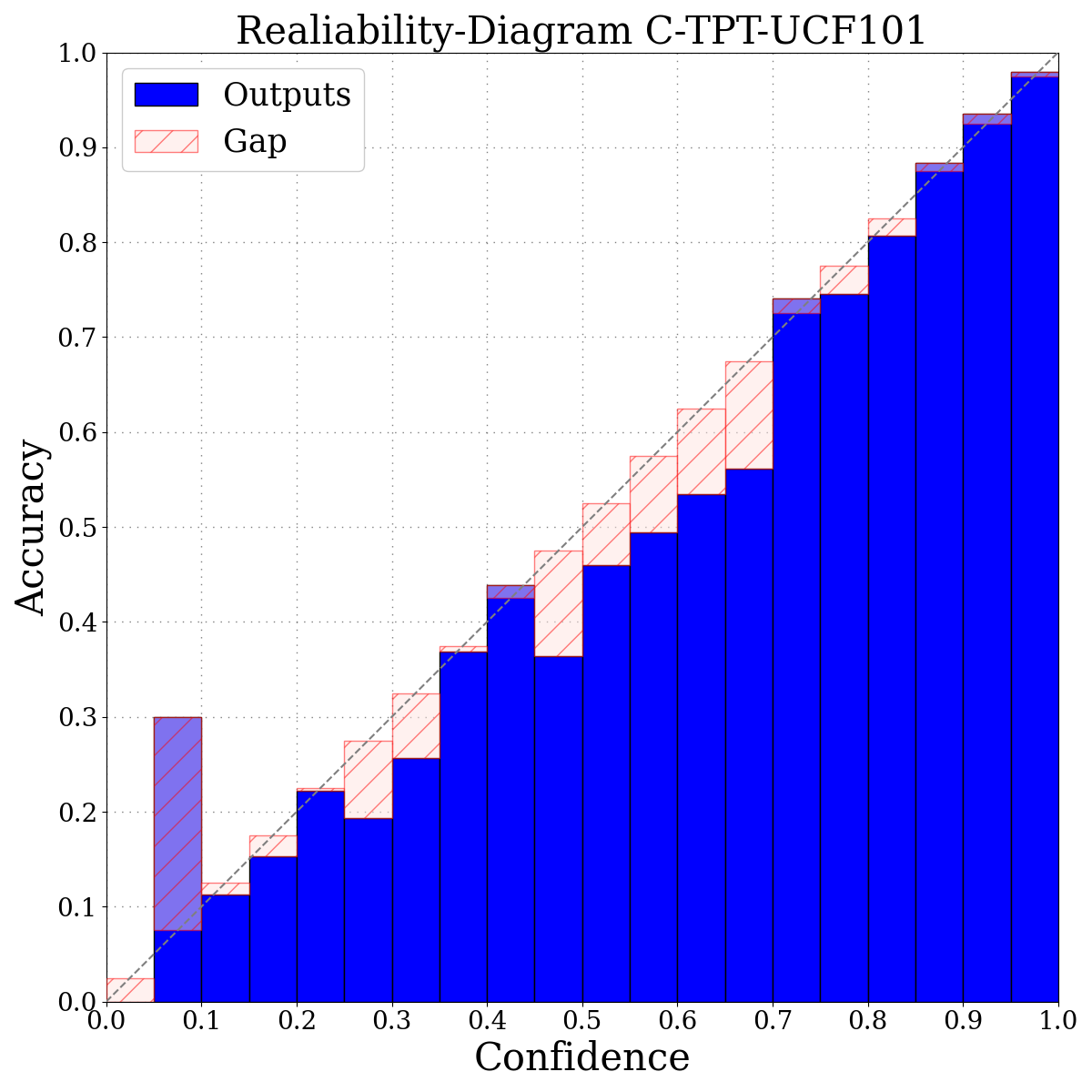}
        \caption{C-TPT:UCF}
        \label{rnctptucf101}
    \end{subfigure}%
    \begin{subfigure}[t]{0.24\linewidth}
        \centering
        \includegraphics[width=\linewidth]{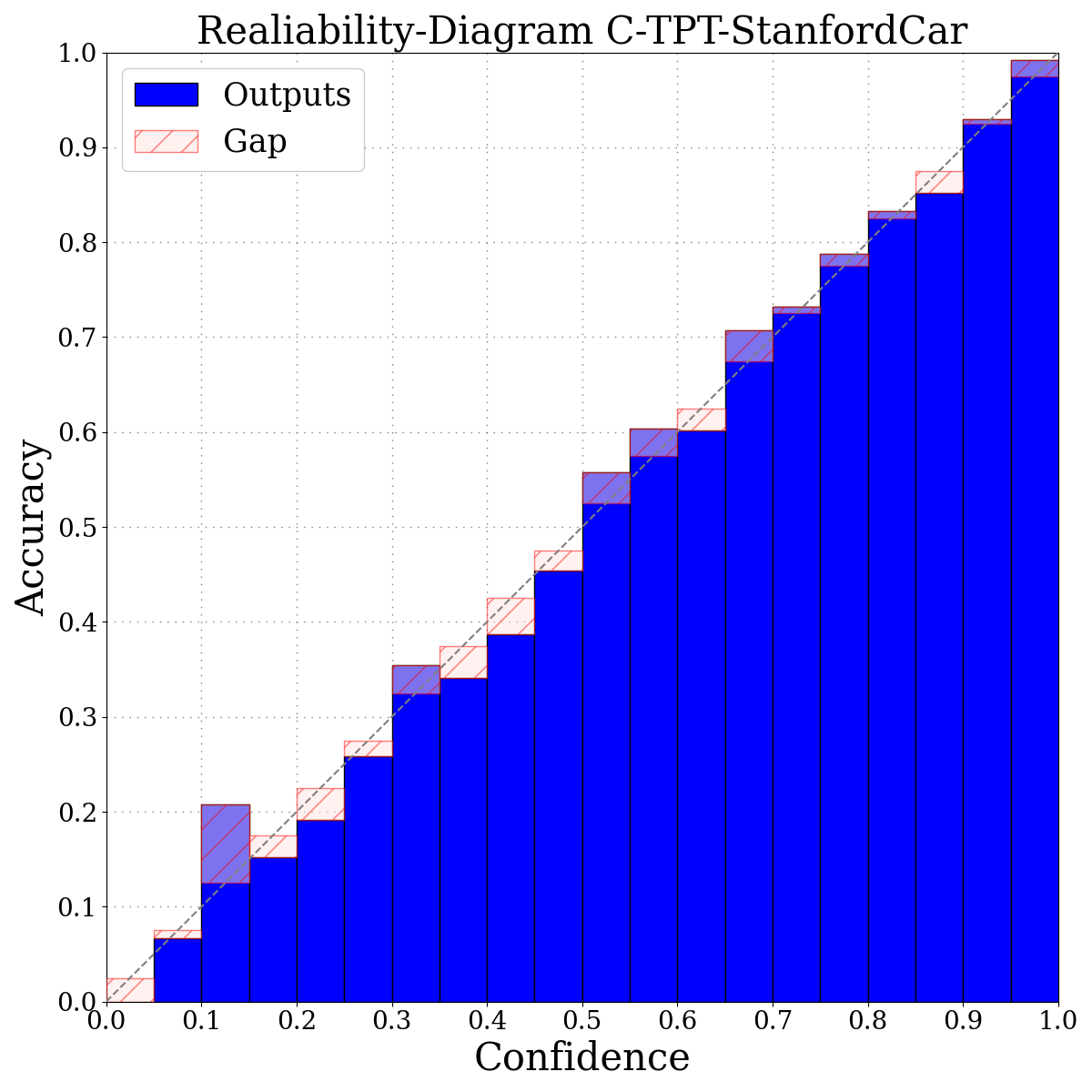}
        \caption{C-TPT:Car}
        \label{rnctpteurosat}
    \end{subfigure}%
    \begin{subfigure}[t]{0.24\linewidth}
        \centering
        \includegraphics[width=\linewidth]{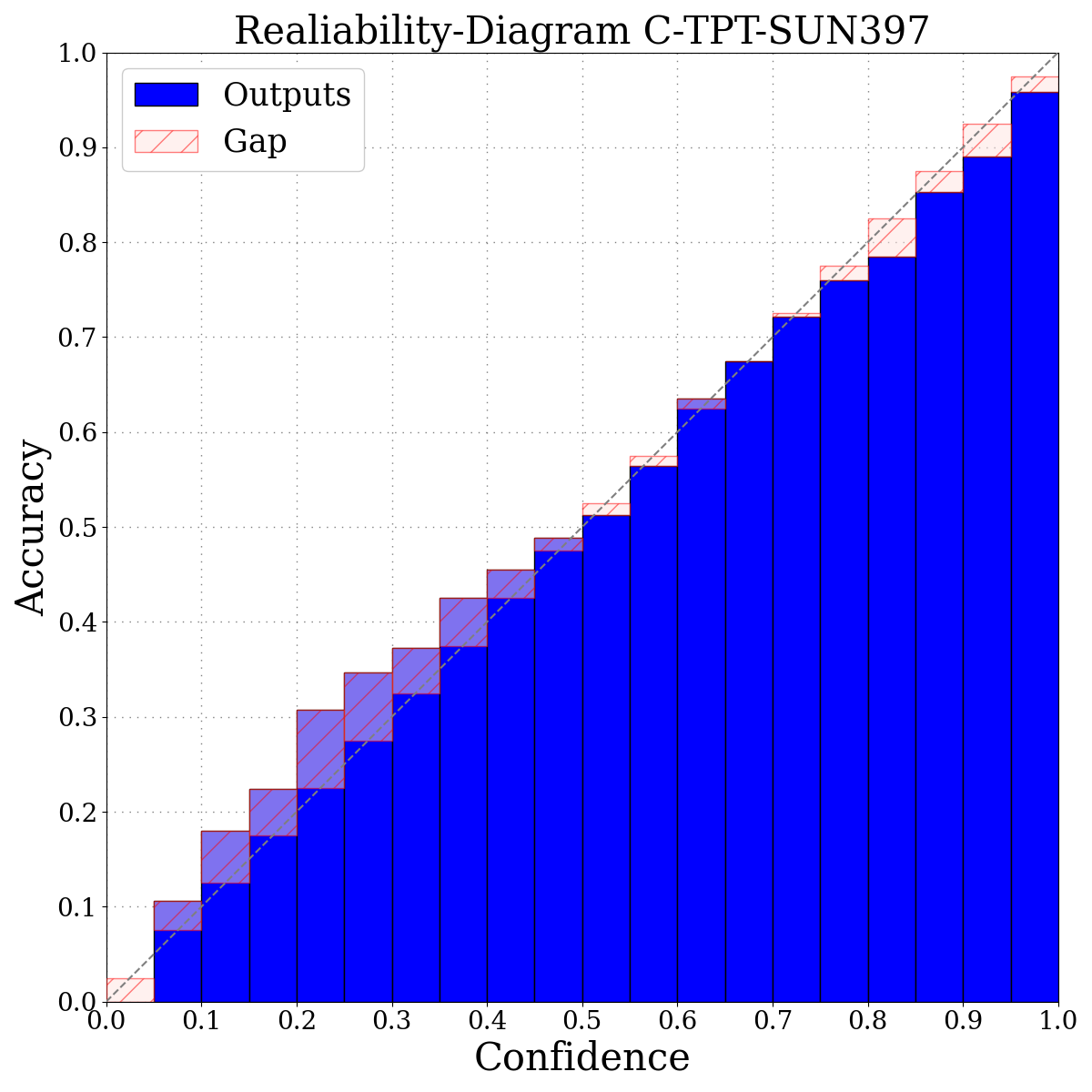}
        \caption{C-TPT:SUN}
        \label{rnctptsun}
    \end{subfigure}%
    
    \vspace{0.5em} 

    \begin{subfigure}[t]{0.24\linewidth}
        \centering
        \includegraphics[width=\linewidth]{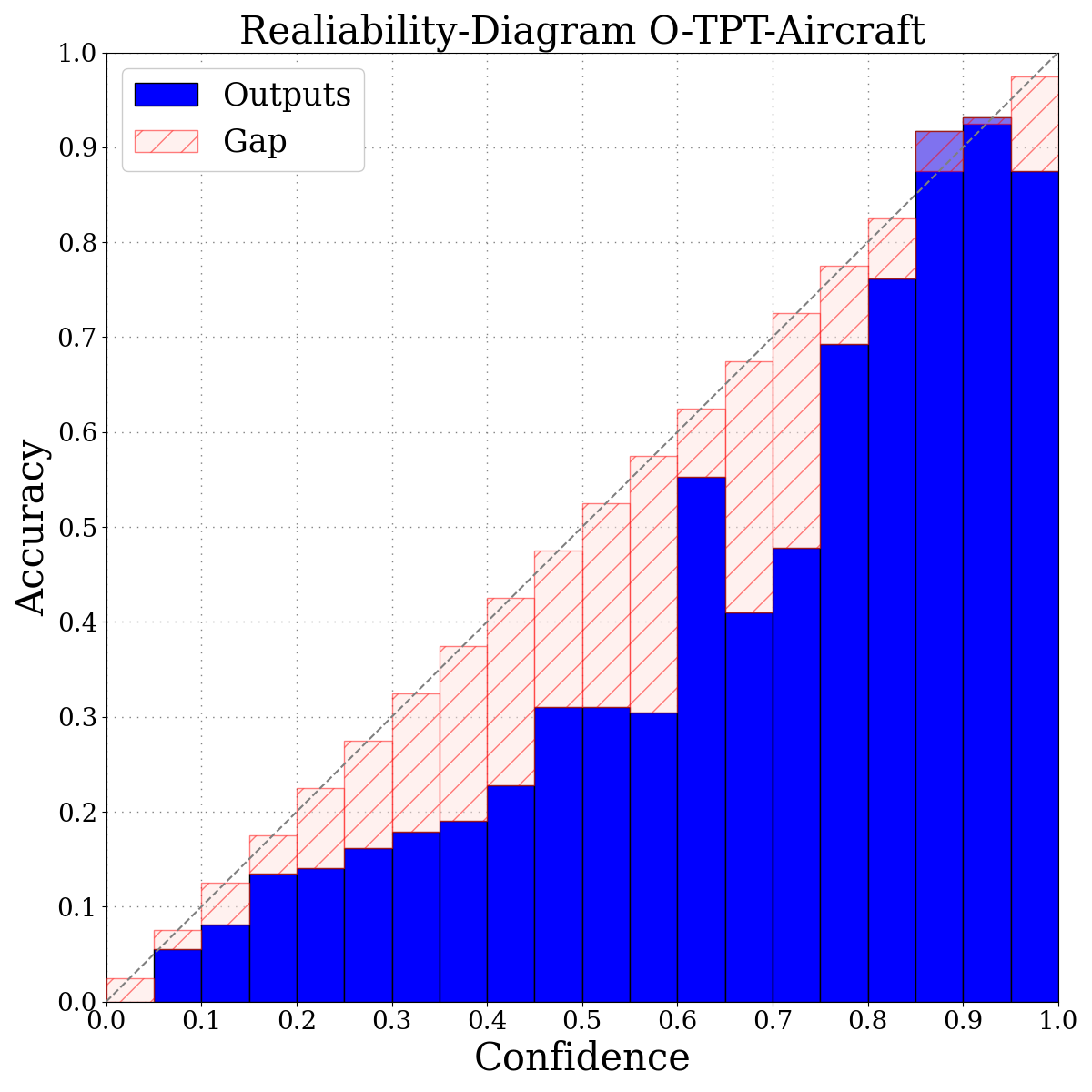}
        \caption{\texttt{O-TPT}:Air}
        \label{rnhtaircraft}
    \end{subfigure}%
    \begin{subfigure}[t]{0.24\linewidth}
        \centering
        \includegraphics[width=\linewidth]{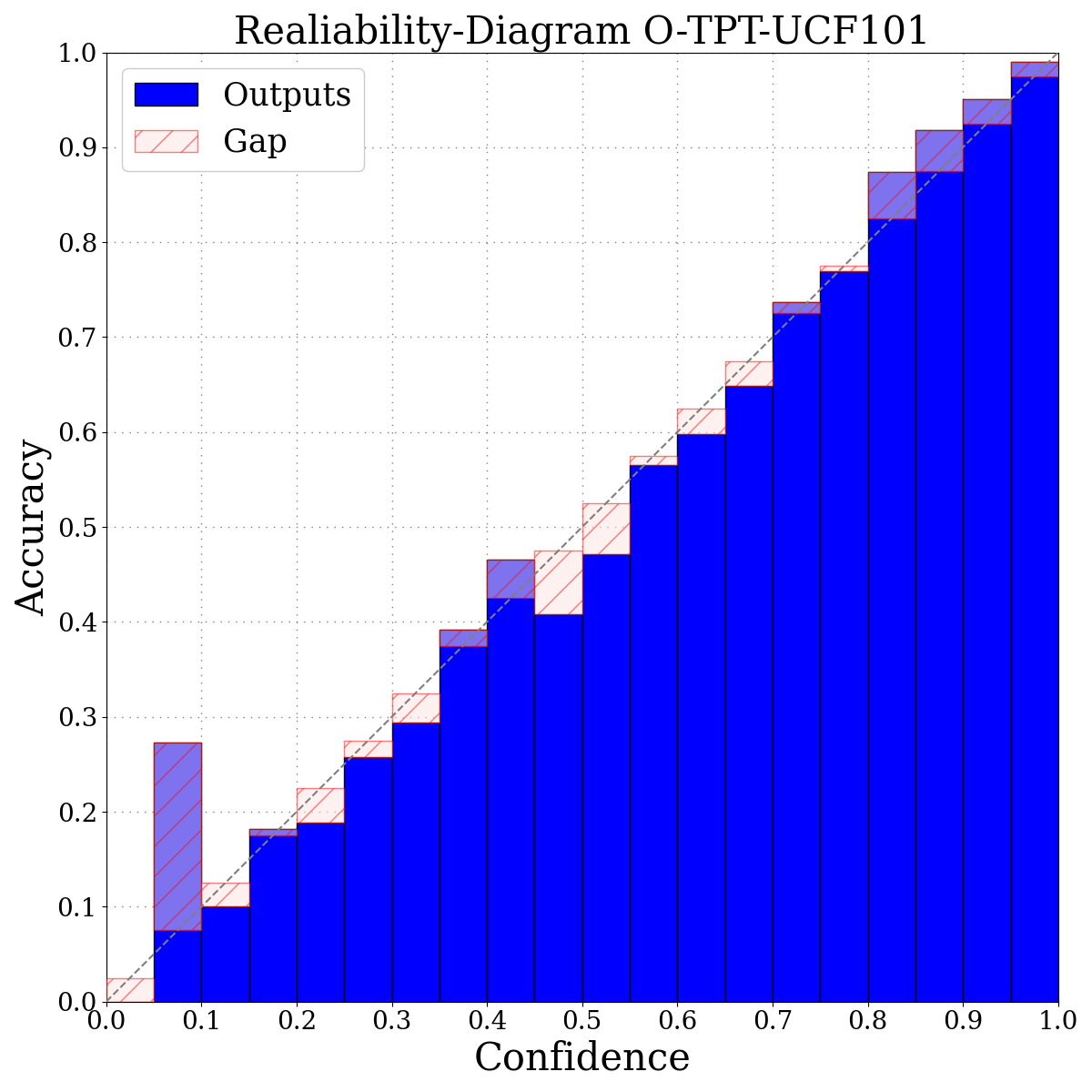}
        \caption{\texttt{O-TPT}:UCF}
        \label{rnhtucf101}
    \end{subfigure}%
    \begin{subfigure}[t]{0.24\linewidth}
        \centering
        \includegraphics[width=\linewidth]{figures/Reliability_diagram-rnC-TPT_18.0_StanfordCar.png}
        \caption{\texttt{O-TPT}:Car}
        \label{rnhteurosat}
    \end{subfigure}%
    \begin{subfigure}[t]{0.24\linewidth}
        \centering
        \includegraphics[width=\linewidth]{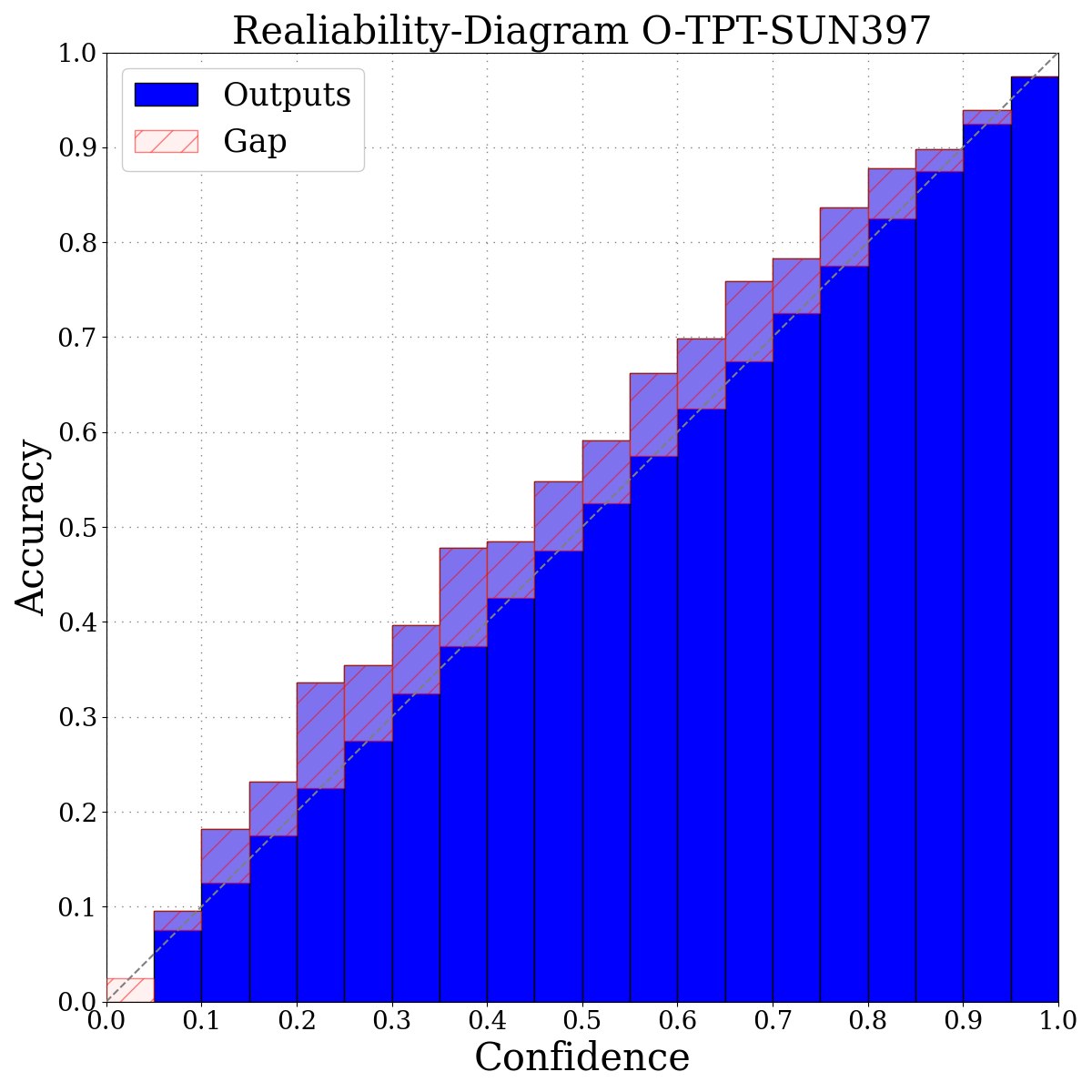}
        \caption{\texttt{O-TPT}:SUN}
        \label{rnhtsun}
    \end{subfigure}%
    
    \caption{Reliability diagrams for CLIP RN-50.}
    \label{fig:RN_reliability_diagram}
\end{figure}

Figure~\ref{fig:vit_reliability_diagram} and Figure~\ref{fig:RN_reliability_diagram} illustrate the reliability diagrams for the CLIP-B/16 and CLIP RN-50 backbones, respectively, comparing the performance of C-TPT and \texttt{O-TPT} across the Aircraft, UCF101, Car, and SUN397 datasets. For the CLIP-B/16 backbone (Fig.~\ref{fig:vit_reliability_diagram}), \texttt{O-TPT} effectively addresses the overconfidence problem and outperforms C-TPT, as evident from the reliability diagrams in the top and bottom rows of Fig.~\ref{fig:vit_reliability_diagram}. Similarly, with CLIP RN-50 backbone, (Fig.~\ref{fig:RN_reliability_diagram}) shows that \texttt{O-TPT} provides significantly better calibration compared to C-TPT, particularly in addressing overconfident predictions.

\begin{table*}[h!]
\centering
\small
\setlength{\tabcolsep}{3pt} 
\renewcommand{\arraystretch}{1.2} 
\definecolor{lightgray}{gray}{0.9}
\scalebox{0.9}{
\begin{tabularx}{\textwidth}{l|c|*{12}{>{\centering\arraybackslash}X}} 
\toprule
\textbf{Method} & \textbf{Metric} & \rotatebox{90}{\textbf{DTD}} & \rotatebox{90}{\textbf{FLW}} & \rotatebox{90}{\textbf{Food}} & \rotatebox{90}{\textbf{SUN}} & \rotatebox{90}{\textbf{Air}} & \rotatebox{90}{\textbf{Pets}} & \rotatebox{90}{\textbf{Calt}} & \rotatebox{90}{\textbf{UCF}} & \rotatebox{90}{\textbf{SAT}} & \rotatebox{90}{\textbf{Car}} & \rotatebox{90}{\textbf{Avg}} \\
\midrule

\multirow{2}{*}{Zero Shot} 
& Acc. & 39.6 & 57.7 & 73.0 & 56.5 & 16.1 & 79.8 & 80.9 & 56.3 & 21.9 & 56.9 & 60.24 \\
& ECE  & 6.94 & 5.14 & 1.49 & 3.33 & 6.42 &  3.30 & 4.79 & 3.76 & 13.9 & 4.83 & 5.39 \\
\midrule

\multirow{2}{*}{TPT} 
& Acc. & 39.2 & 61.6 & 75.8 & 60.2 & 17.4 & 82.6 & 86.5 & 59.7 & 26.3 & 58.8 & 56.81 \\
& ECE & 24.8 & 17.0 & 7.93 & 11.4 & 17.5 & 7.31 & 6.02 & 14.4 & 15.7 & 4.49 & 12.65 \\
\midrule

\multirow{2}{*}{C-TPT} 
& Acc  & 39.1 & 67.0 & 76.0 & 60.3 & 17.4 & 83.5 & 87.1 & 59.6 & 26.1 & 57.2 & 57.33 \\
& ECE  & 18.0 & 6.34 & 3.70 & 8.28 & 13.5 & 1.75 & 2.85 & 8.82 & 11.2 & 1.65 & 7.61 \\

\bottomrule

\multirow{2}{*}{\texttt{O-TPT} (Ours)}  & \ccl Acc.   & \ccl 40.54 & \ccl 65.49 & \ccl 75.51 & \ccl 58.98 & \ccl 15.99 & \ccl 83.78 & \ccl 86.98 & \ccl 58.79 & \ccl 26.89 & \ccl 56.77 & \ccl 56.97 \\
& \ccl ECE & \ccl 12.42 & \ccl 3.03 & \ccl 1.32 & \ccl 3.35 & \ccl 8.36 & \ccl 4.47 & \ccl 3.53 & \ccl 3.27 & \ccl 7.21 & \ccl 2.74 & \ccl \textbf{4.97} \\
\midrule


\end{tabularx}}
\caption{Comparison of calibration performance with CLIP-RN-50 backbone with the prompt of `a photo of the cool \{class\}'}
\label{tab:rnaphotoofthecool}
\end{table*}

\begin{table*}[h!]
\centering
\small
\setlength{\tabcolsep}{3pt} 
\renewcommand{\arraystretch}{1.2} 
\definecolor{lightgray}{gray}{0.9}
\scalebox{0.9}{
\begin{tabularx}{\textwidth}{l|c|*{12}{>{\centering\arraybackslash}X}} 
\toprule
\textbf{Method} & \textbf{Metric}  & \rotatebox{90}{\textbf{DTD}} & \rotatebox{90}{\textbf{FLW}} & \rotatebox{90}{\textbf{Food}} & \rotatebox{90}{\textbf{SUN}} & \rotatebox{90}{\textbf{Air}} & \rotatebox{90}{\textbf{Pets}} & \rotatebox{90}{\textbf{Calt}} & \rotatebox{90}{\textbf{UCF}} & \rotatebox{90}{\textbf{SAT}} & \rotatebox{90}{\textbf{Car}} & \rotatebox{90}{\textbf{Avg}} \\
\midrule

\multirow{2}{*}{Zero Shot} 
& Acc. & 42.4 & 64.7 & 83.9 & 61.4 & 22.3 & 82.5 & 90.9 & 64.8 & 38.8 & 64.6 & 61.63 \\
& ECE & 4.94 & 4.70 & 2.78 & 3.33 & 7.09 &  2.91 & 7.51 & 2.79 & 13.4 & 2.49 & 5.64 \\
\midrule

\multirow{2}{*}{TPT} 
& Acc. & 45.8 & 69.4 & 84.8 & 65.3 & 22.9 & 83.0 & 93.0 & 67.1 & 40.7 & 67.3 & 63.93 \\
& ECE & 20.5 & 12.2 & 5.05 & 7.94 & 16.2 & 7.30 & 2.91 & 11.6 & 20.8 & 6.26 & 11.07 \\
\midrule

\multirow{2}{*}{C-TPT} 
& Acc & 45.4 & 71.5 & 84.3 & 66.0 & 23.6 & 86.9 & 93.8 & 66.4 & 51.5 & 66.6 & 65.6 \\
& ECE & 15.5 & 4.49 & 1.36 & 3.54 & 9.05 & 2.89 & 1.62 & 3.87 & 5.18 & 1.75 & 4..93 \\

\bottomrule

\multirow{2}{*}{\texttt{O-TPT} (Ours)}  & \ccl Acc.   & \ccl 45.45 & \ccl 70.32 & \ccl 84.79 & \ccl 64.5 & \ccl 22.77 & \ccl 87.76 & \ccl 93.35 & \ccl 65.4 & \ccl 51.011 & \ccl 66.25 & \ccl 65.16 \\
& \ccl ECE  & \ccl 11.79 & \ccl 3.22 & \ccl 2.92 & \ccl 4.62 & \ccl 7.92 & \ccl 3.29 & \ccl 3.24 & \ccl 2.63 & \ccl 5.08 & \ccl 1.92 & \ccl \textbf{4.66} \\
\midrule


\end{tabularx}}
\caption{Comparison of calibration performance with CLIP-B/16 backbone with the prompt of `an example of \{class\}'}
\label{tab:vitanexampleof}
\end{table*}

\begin{table*}[h!]
\centering
\small
\setlength{\tabcolsep}{3pt} 
\renewcommand{\arraystretch}{1.2} 
\definecolor{lightgray}{gray}{0.9}
\scalebox{0.9}{
\begin{tabularx}{\textwidth}{l|c|*{12}{>{\centering\arraybackslash}X}} 
\toprule
\textbf{Method} & \textbf{Metric}  & \rotatebox{90}{\textbf{DTD}} & \rotatebox{90}{\textbf{FLW}} & \rotatebox{90}{\textbf{Food}} & \rotatebox{90}{\textbf{SUN}} & \rotatebox{90}{\textbf{Air}} & \rotatebox{90}{\textbf{Pets}} & \rotatebox{90}{\textbf{Calt}} & \rotatebox{90}{\textbf{UCF}} & \rotatebox{90}{\textbf{SAT}} & \rotatebox{90}{\textbf{Car}} & \rotatebox{90}{\textbf{Avg}} \\
\midrule

\multirow{2}{*}{Zero Shot} 
& Acc. & 41.10 & 58.10 & 75.20 & 56.20 & 16.10 & 75.70 & 80.30 & 56.30 & 25.5 & 55.8 & 48.45 \\
& ECE & 5.20 & 3.04 & 3.31 & 3.68 & 4.80 &  2.52 & 7.91 & 3.76 & 9.43 & 4.80 & 4.845 \\
\midrule

\multirow{2}{*}{TPT} 
& Acc. & 41.2 & 62.7 & 76.1 & 60.7 & 17.9 & 77.2 & 87.1 & 57.7 & 29.4 & 57.7 & 56.77 \\
& ECE & 20.2 & 12.2 & 4.83 & 8.19 & 15.2 & 6.98 & 5.12 & 15.3 & 11.1 & 5.52 & 10.46 \\
\midrule

\multirow{2}{*}{C-TPT} 
& Acc & 41.2 & 65.4 & 75.8 & 61.4 & 17.6 & 78.0 & 88.4 & 58.4 & 30.4 & 57.1 & 57.37 \\
& ECE & 15.6 & 2.97 & 1.90 & 4.84 & 7.16 & 2.72 & 2.89 & 6.99 & 7.69 & 2.05 & 5.48 \\

\bottomrule

\multirow{2}{*}{\texttt{O-TPT} (Ours)}  & \ccl Acc.   & \ccl 41.19 & \ccl 65.49 & \ccl 75.62 & \ccl 60.97 & \ccl 16.71 & \ccl 77.79 & \ccl 88.36 & \ccl 57.94 & \ccl 33.32 & \ccl 56.733 & \ccl 57.412 \\
& \ccl ECE  & \ccl 13.59 & \ccl 2.49 & \ccl 1.47 & \ccl 3.38 & \ccl 6.6 & \ccl 2.55 & \ccl 2.56 & \ccl 6.2 & \ccl 5.07 & \ccl 2.69 & \ccl \textbf{4.66} \\
\midrule


\end{tabularx}}
\caption{Comparison of calibration performance with CLIP- RN-50 backbone with the prompt of `an example of \{class\}'}
\label{tab:rnanexampleof}
\end{table*}

\section{Calibration with different prompts} 
\label{sec:diffpromp}
This section presents the results of \texttt{O-TPT} initialized with different prompts, such as `a photo of the cool \{class\}' and `an example of \{class\}', across CLIP-B/16 and CLIP RN-50 backbones. Tables~\ref{tab:vitaphotoofthecool} and \ref{tab:rnaphotoofthecool} summarize the performance of \texttt{O-TPT} with the prompt `a photo of the cool \{class\}' and 5 context token tuning. For CLIP-B/16, \texttt{O-TPT} achieves an overall reduced calibration error of \textbf{5.01}, compared to 7.35 for C-TPT, while for RN-50, it achieves \textbf{4.97}, compared to 7.61 for C-TPT. Similarly, Tables~\ref{tab:vitanexampleof} and \ref{tab:rnanexampleof} present the results for the prompt `an example of \{class\}' and 4 context token tuning. Here, \texttt{O-TPT} again outperforms C-TPT, achieving a reduced calibration error of \textbf{4.66} (CLIP-B/16) compared to 4.93, and \textbf{4.66} (CLIP RN-50) compared to 5.48. These results consistently demonstrate that \texttt{O-TPT} effectively reduces calibration errors across various prompt initializations, showcasing its robustness and adaptability in diverse settings.

\section{Calibration with a Combination of C-TPT and O-TPT}
\label{sec:c-tpt+o-tpt}
Tab.~\ref{tab:otptctpt} shows that O-TPT + C-TPT can outcompete O-TPT in calibration performance, thereby revealing the generalizability of O-TPT over a stronger baseline.

\begin{table}[h] 
    \renewcommand{\arraystretch}{1.2} 
    \setlength{\tabcolsep}{6pt} 
    \begin{tabular}{l|c|c|c|c} 
    \toprule
    \textbf{Method} & \textbf{Metric} & \textbf{DTD} & \textbf{FLW} & \textbf{UCF} \\ 
    \midrule
    \multirow{2}{*}{O-TPT} & ACC & 45.68 & 70.07 & 64.16 \\ 
                            & ECE & 7.88  & 3.87 & 2.34 \\ 
    \midrule
    \multirow{2}{*}{O-TPT + C-TPT} & ACC & 45.2  & 70.6 & 64.1 \\ 
                                    & ECE & \textbf{7.06}  & \textbf{3.41} & \textbf{2.14}  \\ 
    \bottomrule
    \end{tabular}
    \caption{\small O-TPT + C-TPT on DTD,FLW and UCF.}
    \label{tab:otptctpt}
\end{table}
\section{O-TPT results on Medical Prompt tunning methods}

\begin{figure*}[t]
    \centering
    \includegraphics[width=0.9\linewidth]{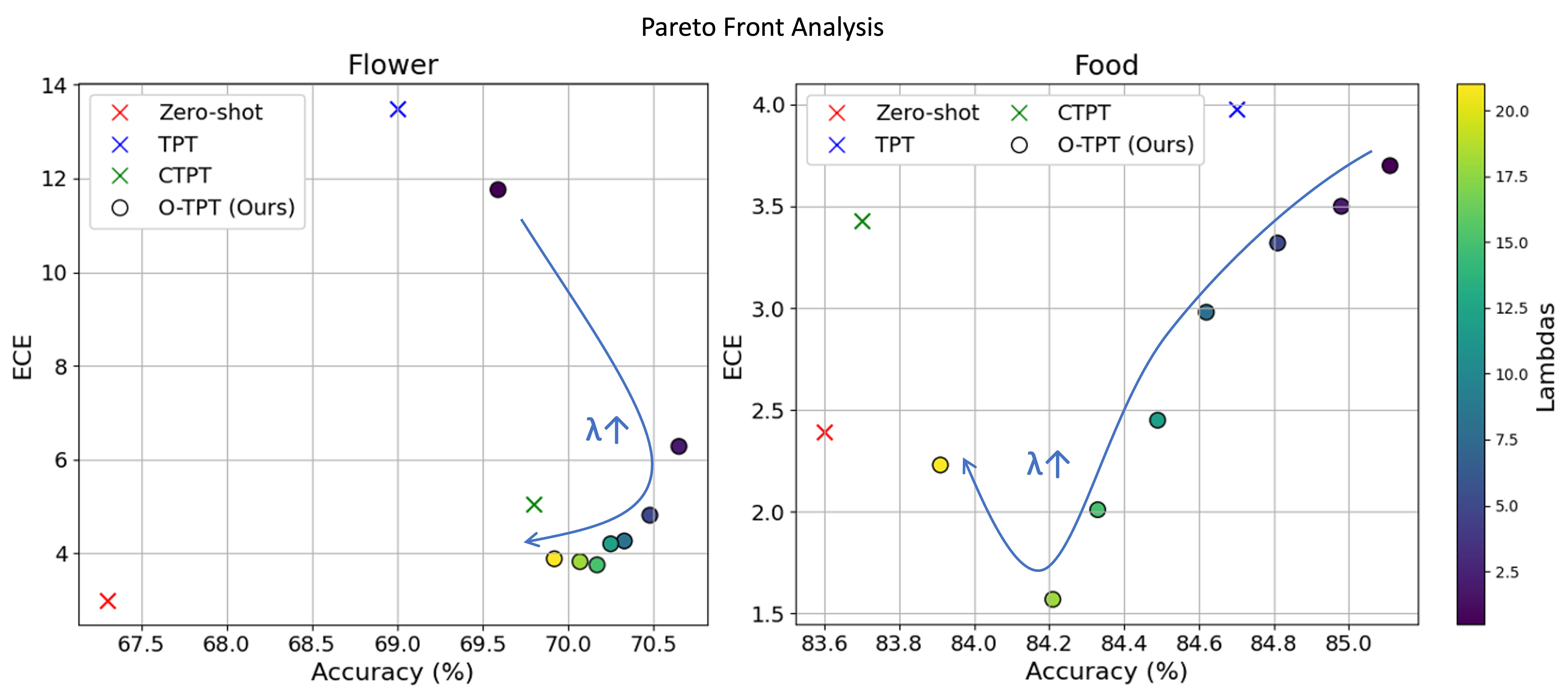}
    \vspace{0.5pt}
    \caption{\footnotesize Pareto front analysis}
    \label{fig:pareto_front}
    \vspace{-8pt}
\end{figure*}

\label{sec:med+o-tpt}
\begin{table}[h] 
    \centering
\renewcommand{\arraystretch}{1.2} 
\setlength{\tabcolsep}{6pt} 
\scalebox{1.0}{
\begin{tabular}{l|c|c|c} 
\toprule
\textbf{Method} & \textbf{Metric} & \textbf{Covid(BA)} & \textbf{Covid(CA)}\\ 
\midrule
\multirow{2}{*}{BAPLe} & ACC & 99.90 & 82.5 \\ 
                        & ECE & 3.21  & 15.64 \\ 
\midrule
\multirow{2}{*}{\begin{tabular}[c]{@{}l@{}}BAPLe+\\ O-TPT\end{tabular}} 
                    & ACC & \text{99.62} & 81.36 \\ 
                    & ECE & \textbf{0.91}  & \textbf{5.97} \\ 
\bottomrule
    \end{tabular}}
    \vspace{-6pt}
    \captionof{table}{MedCLIP: BAPLe + O-TPT on Covid dataset.}
    \label{tab:covidmedclip}
\end{table}

\begin{table}[t]
\centering
\renewcommand{\arraystretch}{1.2} 
\setlength{\tabcolsep}{12pt} 
\scalebox{1}{ 
\begin{tabular}{l|c|c} 
    \toprule
    \textbf{Method} & \textbf{Metric} & \textbf{ISIC'18} \\ 
    \midrule
    \multirow{2}{*}{FPT} & ACC & 98.43 \\ 
                          & ECE & 0.2328 \\ 
    \midrule
    \multirow{2}{*}{\begin{tabular}[c]{@{}l@{}}FPT+\\ O-TPT\end{tabular}} & ACC & 98.25 \\ 
                                & ECE & \textbf{0.1381} \\ 
    \bottomrule
\end{tabular}
}
\caption{ FPT+O-TPT on ISIC2018.}
\label{tab:FPT}
\end{table}

\vspace{10pt} 

\begin{table}[t]

\renewcommand{\arraystretch}{1.2} 
\setlength{\tabcolsep}{12pt} 
\scalebox{1}{ 
\begin{tabular}{l|c|c} 
    \toprule
    \textbf{Method} & \textbf{Metric} & \textbf{KC} \\ 
    \midrule
    \multirow{2}{*}{PS} & ACC & 76.6  \\ 
                          & ECE & 15.54 \\ 
    \midrule
    \multirow{2}{*}{\begin{tabular}[c]{@{}l@{}}PS+\\ O-TPT\end{tabular}} & ACC & 76.2  \\ 
                                & ECE & \textbf{12.73} \\ 
    \bottomrule
\end{tabular}
}
\caption{ PLIP: Promptsmooth (PS)+O-TPT on KatherColon (KC).}
\label{tab:kathercolon}
\end{table}

In Tab.\ref{tab:FPT}, we evaluate FPT\cite{huang2024finegrainedprompttuningparameter} and FPT+O-TPT on ISIC2018, showing encouranging ECE reduction while maintaining accuracy. Tab.\ref{tab:kathercolon} provides comparison using PLIP with Prompt Smooth (PS)\cite{hussein2024promptsmoothcertifyingrobustnessmedical}, where PS+O-TPT improves calibration. Similarly, Tab.\ref{tab:covidmedclip} provides comparison using MedCLIP with BAPLe\cite{hanif2024baplebackdoorattacksmedical}  where BAPLe+O-TPT improves calibration.

\section{Pareto Front analysis with varying lamdas }
\label{sec:parato-analyse}

Fig.~\ref{fig:pareto_front} shows the Pareto front analysis on Food and Flower datasets, highlighting the accuracy-ECE tradeoff with varying $\lambda$s. Our method achieves a better balance than TPT and C-TPT across most $\lambda$ values in two datasets.

\end{document}